%% file: main.tex
\documentclass{bmvc2k}
\def\printOption{arxiv}

\input{packages.tex}

\input{definitions.tex}

\input{frontmatter}

\begin{document}

\maketitle

\def\thefootnote{*}\footnotetext{Equal contribution.}

\input{abstract}

\input{arxiv_addon}

% Paper Part
% ----------

% Introduction
% -------------
\section{Introduction}
\label{sec:introduction}
% FIGURE-1, brief problem def, fails of literature, our motivation,  summary of contributions.

The unprecedented success of convolutional neural networks (CNN) having massive computational and memory complexity has shaped the efforts to find a compromise between the model size and the performance for the effective deployment in devices with limited resources. Knowledge distillation (KD) \cite{Ba_Caruana_NIPS2014, Hinton2015} is a complementary method to those efforts including lightweight model design  \cite{cui2019ArchitectureSearch}, and model compression techniques such as model pruning \cite{Li_2022_CVPR, liu2018rethinking} or quantization \cite{Yamamoto_2021_CVPR}. KD is built on boosting the performance of a relatively smaller model (\ie \emph{student}) by leveraging the knowledge encoded in a powerful model (\ie \emph{teacher}). 

The two critical questions in KD are \emph{how} and \emph{in which form} to transfer the knowledge so that it can benefit the student the most \cite{IJCV_Survey_2021}. Although, the prolific and varied literature of KD includes diverse forms of knowledge to transfer through regressing the predictions \cite{Hinton2015, mirzadeh2020}, the intermediate representations \cite{FitNet,zagoruyko2017paying,Heo2018KnowledgeTV}, and the metrics induced by the distances among the sample representations of either the penultimate layer \cite{tian2019crd,Park_2019_CVPR} or the intermediate layers \cite{Tung_2019_ICCV}; all of these methods have a shared component to facilitate student's learning: regularization with a discrepancy loss between the matching targets.

\input{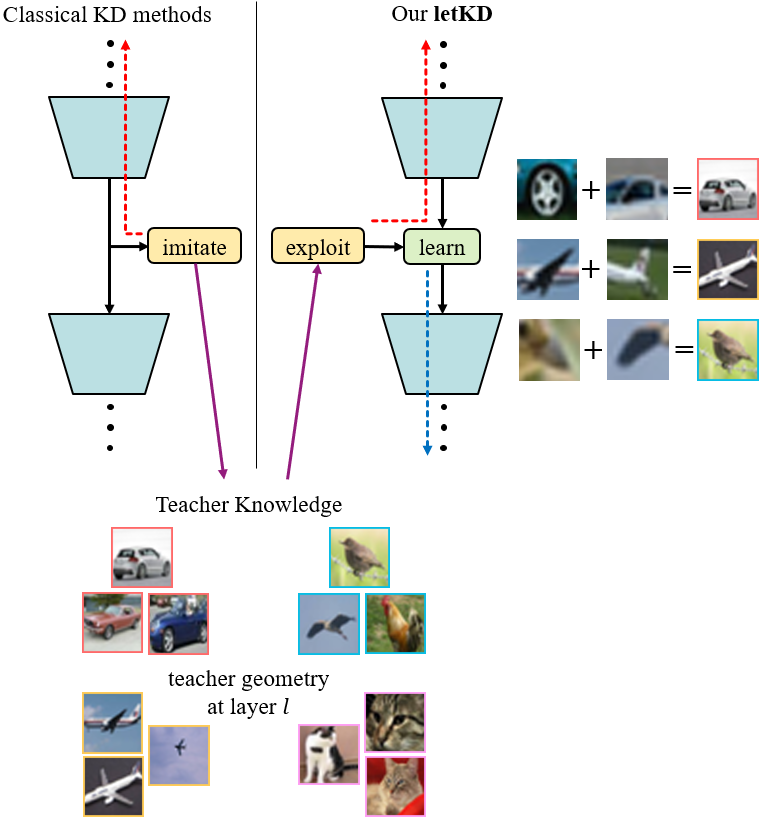}

Common intuition to explain the effectiveness of such regularization is by considering the soft targets learned from a teacher to capture the missing relationships among different categories that sole label supervision cannot provide \cite{Hinton2015}. This intuition can apply well while employing the regularization to the predictions at the penultimate layer since both the student and the teacher share the model beyond, \ie a linear classifier. However, its solidity for the intermediate layers is questionable. Isn't it demanding to force the student to imitate the teacher due to their architectural differences? Surely, there exist methods that propose feature alignment modules \cite{FitNet,Chen_2021_CVPR} to match the feature dimensions between the teacher and the student. Still, the student can fail to exploit those intermediate representations as effectively as the teacher having more layers on top does. Indeed, empirical studies show that knowledge transfer is more effective in the penultimate layer than intermediate layers \cite{jain2019quest}. Although this issue is partially addressed by changing the form of the knowledge into the teacher's coarse decisions which the student can comprehend \cite{Song_2021_CVPR}, its regularization of student's learning is still not that effective. Then, is sole regularization the best choice to transfer knowledge? Moreover, can we explicitly use the teacher's knowledge in the inference as well? In this paper, we try to address those questions within the context of a feature extraction process. We propose a learnable feature transform layer that effectively lets the student decide whether to leverage the teacher's knowledge and use it explicitly during the inference in addition to regularizing the learning. 

Specifically, inspired by \cite{Gorgun_2022_BMVC}, we revamp $1{\sxtimes}1$-BN-ReLU-$1{\sxtimes}1$ based feature transform to assign a feature vector to each local region according to the semantic meaning of the template that the corresponding region matches. We propose to supervise the templates with the semantic entities learned by the teacher and leave the semantic vector learning to the student. This will let the student learn the entities (\eg \emph{wing}, \emph{tire}) that the teacher finds useful and exploit them (\eg \emph{wing} and \emph{beak} $\to$ \emph{bird}) in feature transform, enabling us to feed forward the knowledge rather than imitating it as in \cref{fig:fig1}. To enable learning, we employ a soft-max solution to the best-matching template and represent its feature vector as the weighted combination of the semantic vectors with the matching scores. We rigorously let the solution space include 0-weight, effectively enabling the student to discard transferred knowledge. Hence, the student is not only able to reshape the transferred knowledge with its semantic vectors but also feed forward it to the upper layers, which is a novel approach in KD.

To validate our claims, we design an extensive empirical study. The results confirm that feed forwarding the teacher's knowledge by explicitly using it in feature transform improves student models, and our layer enables the utilization of the teacher's knowledge during inference. We tested our method on 10 student models and 3 classification benchmarks, showing its wide applicability. We set a new state-of-the-art by consistently improving upon the direct application of multi-layer teacher supervision \cite{jain2019quest, Song_2021_CVPR} and other KD methods in both single and multi-layer transfer settings.

%through our layer, the teacher's knowledge can be kept in memory to be used during inference as well. 

%To validate our claims, we conducted a thorough empirical study. The results confirm that utilizing the teacher's knowledge in feature transformation leads to improved student models. Additionally, our approach allows the teacher's knowledge to be stored and utilized during inference.

%In our extensive empirical study, we confirmed that leveraging the teacher's knowledge in feature transformation improves student models. Furthermore, our approach enables the retention and utilization of the teacher's knowledge during inference.

%We applied our method on 10 different student architectures and tested on 3 classification benchmarks. Setting new state-of-the-art, the results show consistent improvements with respect to direct application of multi-layer teacher supervision \cite{jain2019quest, Song_2021_CVPR} as well as other KD alternatives in both single and multi-layer knowledge transfer settings.

%Moreover, our KD layer can be used in any architecture.

\section{Related Work}
\label{sec:review}
%kernel matching based is similar but ours is different, we employ the behaviour decision process to guide the fearure extraction
%they do not use teacher info in feature extraction process.
%in this work we propose to use this supervision explicitly in feature extraction process.

\textbf{Our contributions.} Prior to discussing the works that are most related to ours, we recapitulate our contributions as $i)$ we propose a learnable KD layer that captures the teacher's knowledge during training and employs it in feature transform, effectively feeding forward the transferred knowledge deeper and enjoying it during the inference as well, $ii)$ we repurpose some convolution kernels of the teacher as the cluster centers to semantic entities and exploit them in KD, $iii)$ we introduce a novel form of supervision based on the teacher's decisions on the intermediate layers.

\textbf{KD in penultimate layer.} Leading momentum in KD is built on transferring the inter-category relations captured by the teacher. Thus, many works regularize the student's learning by matching its predictions with the teacher's soft predictions \cite{Hinton2015,mirzadeh2020,xu2020kd,Zhao_2022_CVPR}. Following a similar perspective, relations among the local features are also exploited as the teacher's knowledge, which include the metrics induced by the distances among the sample representations \cite{tian2019crd,Park_2019_CVPR,Ye_2020_CVPR} and fine-grained labeling obtained by clustering the teacher's features \cite{jain2019quest}. KD regularization in the penultimate layer typically outperforms its intermediate layer counterparts since both the teacher and the student have the same representation power on top, \ie a linear classifier. Indeed, recent work \cite{Chen_2022_CVPR} shows that the student can directly copy the teacher's classifier once their features match. In our work, we also build on teacher's soft labeling for KD. Differently, we do not employ it as a regularizer. We instead store them within the parameters of our KD layer and exploit them in feature transform.
%Distilling task knowledge via relationshop matching, CVPR 2020: \cite{Ye_2020_CVPR} relation based,
%it first distills the discriminative embedding by aligning triplets, e.g. the relative similarities between two impostors are specified by the teacher; REFILLED then distills the classification ability via local embedding-based classifiers.
%Correlation congruence for KD, ICCV 2019, Peng et al. \cite{Peng_2019_ICCV}: relation based, ijcv surveyde hint layer olarak yazıyor

\textbf{KD in intermediate layers.} Lacking category-based annotations from the teacher, KD at lower layers uses other forms of supervision including local features \cite{FitNet,Heo_2019_ICCV}, saliency maps \cite{zagoruyko2017paying}, feature distributions \cite{Ahn_2019_CVPR,Chen_2021_CVPR}, and the metrics induced by the inter-feature distances \cite{Tung_2019_ICCV,Peng_2019_ICCV}. Such methods push the student to imitate the geometry of the teacher's intermediate representations. However, the intuition of capturing such relations does not directly apply in the intermediate layers as it does in the penultimate layer. The student can fail to exploit those intermediate representations as effectively as the teacher having more layers on top does. Although recent works explicitly study mitigating this problem by supervising penultimate layer using multiple intermediate layers with improved feature alignment modules \cite{Chen_2021_CVPR}, selectively deciding which intermediate layers to distill \cite{Zhu2021} or changing the form of the knowledge into the teacher's coarse decisions \cite{Song_2021_CVPR}, it is still empirically observed that including KD in intermediate layers has a negative effect \cite{jain2019quest}. In contrast to existing efforts, we profitably exploit the teacher's knowledge in the intermediate layer with our KD layer. Our approach enables the student to build new representations from the semantic entities learned by the teacher as well as to discard the nuisance information, which differs from existing selective feature distillation schemes \cite{Yang2022}. 

\textbf{Deeply supervised nets.} Our student's learning scheme is closely related to the methods that use auxiliary classification loss to regularize the features and to facilitate learning without vanishing gradients \cite{szegedy2015going, lee2015deeply}. Differently, we explicitly use such intermediate predictions to semantically represent local regions with the combination of learned vectors weighted by those predictions. Similar to us, class-level predictions are used to shape the behavior of the intermediate features in \cite{Gorgun_2022_BMVC} for the classification problem. Different from them, we relate such a mechanism to KD for the first time and we additionally propose a novel form of supervision that is based on the local decisions of the teacher, yielding effective sub-class annotations.

% ADA NOTE:
% Preliminary 4'te başlayabilir mi?

% PROBLEM DEFINITION
% -------------------
\section{Preliminaries}
\label{sec:notations}
Consider the mapping $f(\cdot;\theta):\mathcal{X}\rightarrow \mathcal{Y}$ within $L$-layer composite function family, \ie $f=f_{L} \circ f_{L\shortminus1} \circ \cdots f_2 \circ f_1$, where $\mathcal{X}$ is the space of data points, $\mathcal{Y}$ is the space of labels, and $\theta$ is the model parameters. We consider two models, \ie the student $f_s(x;\theta_s)$, and the teacher $f_t(x; \theta_t)$, with $\vert \theta_t \vert> \vert \theta_s \vert$. Given samples $\{x_i\}_i
{\sim} \mathcal{X}$, we let $\mathcal{S}_l=\{f_s^{(l)}(x_i;\theta_s)\}_i$ and $\mathcal{T}_{l^\prime}=\{f_t^{(l^\prime)}(x_i;\theta_t)\}_i$ denote the set of student features and the teacher features at layer $l$ and $l^\prime$, respectively with $f^{(l)}=f_{l} \circ f_{l\shortminus1} \circ \cdots \circ f_2 \circ f_1$. We consider the following matching cost:
\begin{equation}\label{eq:kd_loss}
    \mathcal{L}_{KD}(\mathcal{S}_l,\mathcal{T}_{l^\prime}) = \Vert g_s(\mathcal{S}_l) - g_t(\mathcal{T}_{l^\prime}) \Vert_{\mathcal{M}}
\end{equation}
where $\Vert \cdot_1 \shortminus \cdot_2\Vert_{\mathcal{M}}$ is a metric to compare its arguments, $g_s$ and $g_t$ are the transformations to match the dimensions of $\mathcal{S}_l$ and $\mathcal{T}_{l^\prime}$. For example, when $\mathcal{S}_l$ and $\mathcal{T}_{l^\prime}$ correspond to model predictions \cite{Hinton2015}, $g_s$ and $g_t$ are \emph{scaled soft-max}, and $\mathcal{M}$ is \emph{KL-div}. %as in the seminal work

Given the samples $\{(x_i,y_i)\}_i{\sim} \mathcal{X}{\sxtimes}\mathcal{Y}$ for the classification task, typical KD methods regularize the student's learning with $\mathcal{L}_{KD}$ to transfer the teacher's knowledge as:
\begin{equation}\label{eq:total_loss_with_kd}
    \mathcal{L}(\theta_s) = \mathcal{L}_{CE}(\{(f_s(x_i;\theta_s),y_i)\}_i) + \lambda \mathcal{L}_{KD}(\mathcal{S}_l,\mathcal{T}_{l^\prime})
\end{equation}
where $\mathcal{L}_{CE}$ is the \emph{cross-entropy} loss and $\lambda$ is the weight of the distillation loss. In the rest of the paper, we additionally consider employing the teacher's knowledge in feature transform as $f=\cdots f_{l+1}\circ f_{KD} \circ  f_{l}\cdots$ with $f_{KD}$ as a function of $\mathcal{T}_{l^\prime}$ and propose a learnable layer to use it without teacher's feedback $\mathcal{T}_{l^\prime}$ in the inference.

% METHOD
% -------
\section{Method}
\label{sec:method}
We propose a lightweight residual layer with $1{\sxtimes}1$-BN-ReLU-$1{\sxtimes}1$ convolution block for KD. Our layer enhances its input feature map with the knowledge transferred from the teacher. We first explain our theoretical motivation in \cref{sec:kdlayer_formulation}. The theory suggests that we can use the teacher's knowledge rather explicitly in feature extraction and feed forward it deeper if the kernels of the first $1{\sxtimes}1$ convolutions are guided by the teacher's supervision. We use our layer in both the penultimate layer and an intermediate layer. To facilitate learning, we propose different teacher supervision mechanisms for both in \cref{sec:labeling}, where we propose a novel form of supervision that is based on the teacher's decisions.

\input{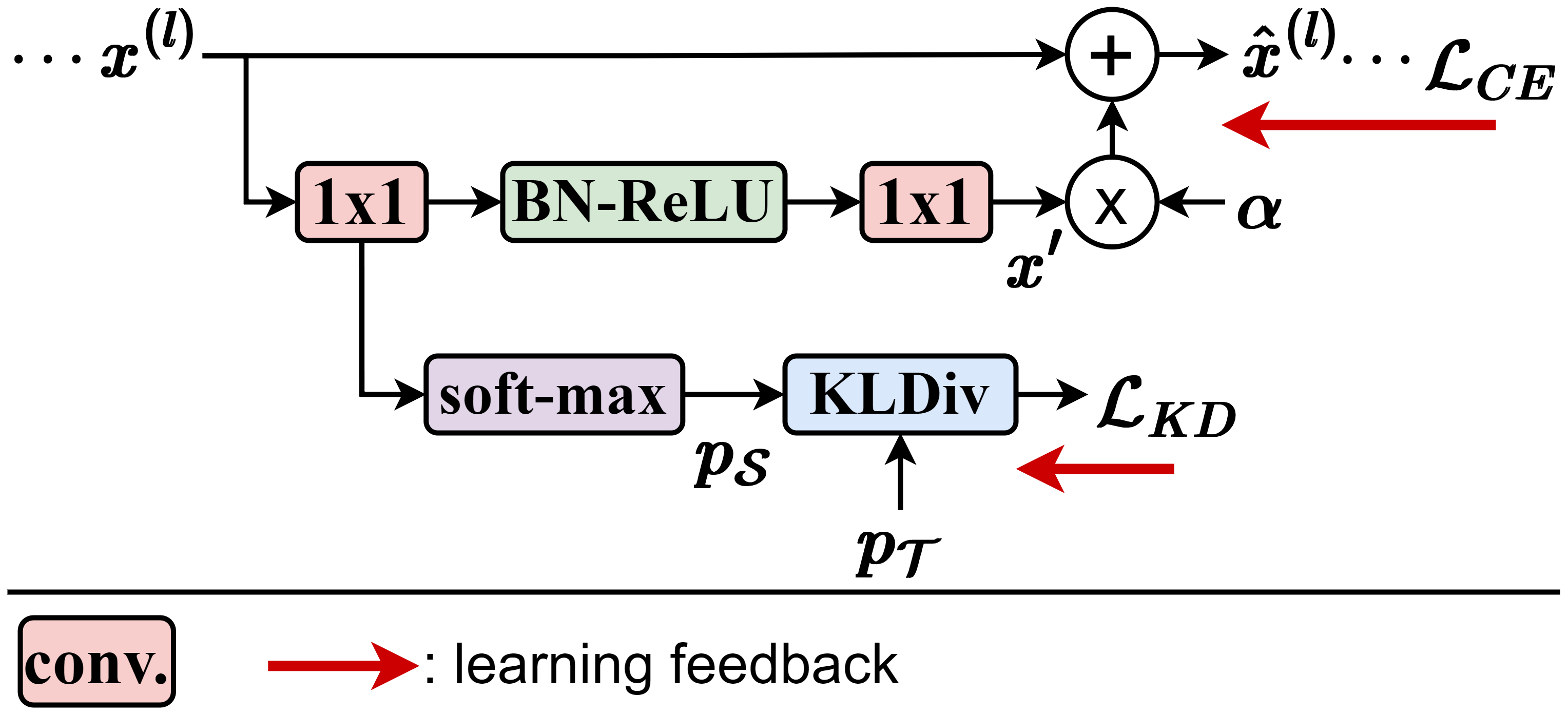}

\subsection{Learnable KD Layer}
\label{sec:kdlayer_formulation}
Our KD technique is built on the perspective \citep[and references therein]{gurbuz2023generalizable,gurbuz2019novel,zhou2018interpreting,zeiler2014visualizing}, explaining the success of the CNNs by considering each pixel of a feature map as a semantic entity (\eg \emph{wing}, \emph{tire}, \emph{etc.}). We aim to transfer such semantic entities learned by the teacher to the student. Let $x^{(l)}$ denote a $h \sxtimes w$ feature map after the $l^{th}$ layer and $x^{(l)}_i{\in}\R^d$ is its $i^{th}$ spatial feature. Our KD layer depicted in \cref{fig:kdlayer} transforms the feature at each pixel-$i$ as:
\begin{equation}\label{eq:kd_layer}
    \hat{x}^{(l)}_i = x^{(l)}_i + \alpha x^\prime_i
\end{equation}
where $x^\prime_i = g(x^{(l)})_i $ is the $i^\text{th}$ spatial feature of the map after we apply $g$, and $\alpha>0$ is a scalar constant. We implement $g$ as $1{\sxtimes}1$-BN-ReLU-$1{\sxtimes}1$ where $1{\sxtimes}1$ is convolution with unit spatial extent, \ie linear transform. We leverage the teacher's knowledge to supervise the learning of the first convolution kernels. We now technically explain how such a simple residual block is an effective way for knowledge transfer.

\textbf{Motivation.} We first repurpose $g\coloneqq 1{\sxtimes}1$-BN-ReLU-$1{\sxtimes}1$ transform as feature embedding by template matching, following \cite{Gorgun_2022_BMVC}. Specifically, we first show that $g$ selects the best-matching kernel to $x^{(l)}_i$ and assigns a feature vector to pixel-$i$ according to the semantic meaning of the matched kernel. We then regularize the learning of the matching kernels by predicting the fine-grained labels (\eg \emph{wing}, \emph{tire}, \emph{etc.}) provided by the teacher. Namely, the matching kernels become the weights of a linear classifier (\ie class representatives). In the end, the predictions match the teacher's decisions for local regions and the student assigns semantic vectors based on those predictions. Hence, the student controls how to use the transferred knowledge. Moreover, the knowledge is explicitly embedded in the feature transform, effectively enabling its use during inference.

\textbf{Formulation.} For CNNs, $x^{(l)}_i$ represents a local region around it to some spatial extent depending on the depth. To simplify notation, we use $x_i=x^{(l)}_i$ henceforth. We consider a set of matching kernels $\{\omega_k\in\R^{d}\}_k$ as templates, each of which seeks for a particular pattern. To each kernel $\omega_k$, we associate an embedding vector $\nu_k\in\R^{d^\prime}$ representing the semantics of the corresponding pattern. We aim to replace $x_i$ with the embedding vector of its best-matching kernel. We formally write this process as:
\begin{equation}\label{eq:argmax}
p_{\mid i} =  \argmax_{p,q\geqslant 0} q\,\mu+\Sigma_k p_k\, \omega_k\T x_i \quad \text{s.to} \quad q+\Sigma_k p_k =1
\end{equation}
where $\mu$ is a threshold enabling to zero out the embedding vector if no kernel matches with at least $\mu$ similarity. Then, we assign the representation of $x_i$ as $x^{\prime}_i = \Sigma_k p_{k\mid i}\,\nu_k$ since $p_{\mid i}$ is either one-hot or zero vector owing to \textit{total unimodularity} \cite{unimodular}. To enjoy analytical gradients, we employ entropy smoothing to the objective in \eqref{eq:argmax} as:
\begin{equation}\label{eq:entropy_argmax}
p_{\mid i} = \argmax_{p,q\geqslant0} q\,\mu+ p\T a_{\mid i} -\tfrac{1}{\epsilon}(q\log q+p\T\log p) \quad \text{s.to} \quad q+\Sigma_k p_k =1
\end{equation}
and obtain a soft-max solution $p_{k\mid i} = \tfrac{\exp(\epsilon a_{k\mid i})}{\exp(\epsilon \mu)+\Sigma_{k^\prime}\exp(\epsilon a_{k^\prime\mid i})}$ where $a_{k\mid i} = \omega_k\T x_i $ and $\epsilon$ controls the smoothness of $p_{\mid i}$. Hence, we can implement the feature embedding by template matching via $1{\sxtimes}1$-SoftMax-$1{\sxtimes}1$ with $\{\omega_k\}_k$ and $\{\nu_k\}_k$ as the convolution kernels. Finally, when we obtain this smooth labeling from the teacher as $p_{\mathcal{T}}(i)$ for each pixel-$i$, we can regularize the learning of $\{\omega_k\}_k$ by minimizing the \emph{KL-div} between the predictions of the student $p_{\mathcal{S}}(i)\coloneqq\mathrm{softmax}(a_{\mid i})$ and  $p_{\mathcal{T}}(i)$ for all pixels as $\mathcal{L}_{KD}$ in \eqref{eq:kd_loss} with $\lambda=1$.

% ADA NOTE:
% Practical simplifications detayı supplementarye yönlendirsek? Başka makalelerde de gördüm yapıldığını, çünkü zaten BMVCde göstermiş oluyoruz.

\textbf{Practical simplifications.} With the smooth assignments obtained by \eqref{eq:entropy_argmax}, the student shapes the teacher's knowledge by the weighted combination of its embedding vectors $\{\nu_k\}_k$ with the weights proportional to the matching scores. To enable the student to discard nuisance information from the teacher, we must set a proper $\mu$. As empirically validated in \cite{Gorgun_2022_BMVC} as well as formally discussed in \refappendix, we can indeed inherently learn it from batch statistics by replacing soft-max solution of $p_{k\mid i}$ in \eqref{eq:entropy_argmax} with $\mathrm{BN}$-$\mathrm{ReLU}$, \ie $p_{\mid i}\approx\mathrm{BN}\text{-}\mathrm{ReLU}(a_{\mid i})$.

% Formally, batch normalization (BN) \cite{normalization2015accelerating} performs activity normalization of the form $\hat{a}_k=\gamma_k\tfrac{a_k-\mathbb{E}[a_k]}{\sqrt{\mathrm{Var}(a_k)}}+\beta_k$, \ie learnable affine transform upon mapping around 0. Thus, the unnormalized assignment weights after BN approximately have the form $\hat{\gamma} u+\hat{\beta}$ since $\mathrm{e}^u \approx 1 + u$ around 0. Hence, we can simply employ $\hat{p}_k=\mathrm{max}\{0, \hat{a}_k\}$, \ie $\mathrm{ReLU}$, to zero-out the assignment vector since BN will learn proper parameters, $(\beta_k,\gamma_k)$, using the batch statistics to assess the poor matching scores. For the pixels with non-zero activations after BN-ReLU, we can obtain the normalized assignment vector as $\nicefrac{\hat{p}_k}{\eta}$ where $\eta\coloneqq\Sigma_k\hat{p}_k$. That said, we empirically find absorbing $\eta$ into $\alpha$ in \cref{eq:kd_layer} is useful to adaptively put more emphasis on the teacher knowledge according to the matching scores.

% for a derivation, see supplement
% (Sec. D).

%\textbf{Learning.} put finally paragraph etc

\subsection{Teacher Supervision}
\label{sec:labeling}

A critical desiderata of our KD layer is per pixel label annotations, $p_{\mathcal{T}}{(i)}$, provided by the teacher. We propose different forms of teacher supervision for the uses in the penultimate layer and the intermediate layers. In the following, we assume that the spatial dimensions of the teacher's and the student's feature maps match at the layers where the knowledge transfer is performed, noting that we can transform the student's feature map with \emph{pooling, strided convolution, etc.} to match the dimensions.
 
\subsubsection{Penultimate Layer}
\label{sec:pen_labelling}

Expanding on the idea \citep{gurbuz2023generalizable} that we view each feature map pixel as a semantic entity, we propose to employ $K$-means to the teacher's features at the penultimate layer to obtain fine-grained labels for the semantic entities. We then annotate each pixel by soft-max assignments to the cluster centers to capture inter-category relations.

%%%%%%%%%%%%%%%%%%% x_i notasyonları karışıyor, bir yerde sample image olarak kullanıyoruz, bir yerde de i-th pixel olarak kullanıyoruz

Formally, given the dataset samples $\{x_i\}_i
{\sim} \mathcal{X}$ (or a subset of it), we compute the teacher's feature maps at the penultimate layer, \ie $\{f_t^{(\shortminus 1)}(x_i;\theta_t)\in\R^{h{\sxtimes}w{\sxtimes}d}\}_i$. Considering each pixel as a feature sample, we fit $K$-means clustering to the pixels of those maps and obtain the centers $\{\rho_k\}_{k\in[K]}$ where $[K]=1,\ldots,K$. During training, we pass the input, $x$, through the teacher to obtain $f_t^{(\shortminus 1)}(x;\theta_t)$. We then compute the distance $d_{k\mid i} = \Vert f_t^{(\shortminus 1)}(x;\theta_t)_i - \rho_k  \Vert_2^2$ for each pixel-$i$ of the feature map to the cluster centers and obtain their $K$-dimensional soft labeling as $p_{\mathcal{T}}{(i)}=\mathrm{softmax}(d_{\mid i})$. Although this form of supervision is previously applied to KD \cite{jain2019quest} to match the responses as in \eqref{eq:kd_loss}, we differently repurpose it as a supervision for our matching kernels and enable the student to rather exploit it in feature transform than imitate through our KD layer (\cref{fig:kdlayer}). We summarize our approach in the supplementary material \refalgos.

\textbf{Without $K$-means.} We can directly obtain the soft assignments to the semantic entities from the architectures (\eg ResNet \cite{HeResnetv1_2015,he2016identity}) that involve similar blocks to $3{\sxtimes}3$-BN-ReLU-$1{\sxtimes}1$ by design. Building on our analysis in \cref{sec:kdlayer_formulation}, kernels of $3{\sxtimes}3$ correspond to learnable templates (\ie cluster centers) of some semantic entities. Hence, we can use the soft-maxed activations of the $3{\sxtimes}3$ convolution at the final block to obtain $p_{\mathcal{T}}$. Supporting our claims in \cref{sec:kdlayer_formulation}, we empirically show in \cref{tab:3x3_vs_quest} that this clustering-free approach has competitive performance with much fewer centers compared to $K$-means. 

\subsubsection{Intermediate Layer}
\label{sec:int_labelling}

Granted that we can apply $K$-means based supervision in \cref{sec:pen_labelling} to intermediate layers as well, its data-driven annotation mechanism corresponds to representing the geometry of the teacher's features at some point. {Although this is beneficial in the penultimate layer due to the
shared classifier architecture afterward, it cannot be effectively used by the student's intermediate layers due to capacity differences}. In fact, we empirically observe in \cref{tab:quest_multi} that such supervision without our KD layer is detrimental to the performance. Although our KD layer is a remedy thanks to not pushing the student to imitate but to exploit the knowledge, it can benefit more from another form of supervision. In particular, decision-based supervision \cite{Song_2021_CVPR} has recently been shown to be superior to representation-based, yet it is tailored for coarse decisions of the teacher at the intermediate layers. Thus, as summarized in the supplementary material \refinteralgo  as well as visualized in \cref{fig:intermediate_kd}, we now propose a new supervision based on localized fine-grained decisions to facilitate the learning of our KD layer.

To transfer teacher's localized decisions, we enhance $K$-means based annotations by exploiting the original labels of the image. Specifically, we compute the teacher maps at layer-$l^\prime$, \ie $\{f_t^{(l^\prime)}(x_i;\theta_t)\in\R^{h{\sxtimes}w{\sxtimes}d}\}_i$, for the samples $\{(x_i,y_i)\}_i$ in the dataset (or a subset of it). Considering each pixel of $f_t^{(l^\prime)}(x_i;\theta_t)$ as a feature sample with class label $y_i$, we fit a linear classifier using linear discriminant analysis (LDA) to set the stage for the rest of the formulation. Once fitted, LDA is simply a $1{\sxtimes}1$ convolution (\ie per pixel linear transform) that aims to map the features close only if they share the same label, reflecting localized decision capacity of the teacher at layer-$l^\prime$.
%Surely, we expect some features to be shared among different classes (\eg \emph{tire} for \emph{truck} and \emph{car}) while some are discriminative (\eg \emph{beak} for \emph{bird}). To capture such localized decisions, we apply $K$-means of small $K$ to the features of each class separately and obtain $K$-many sub-class centers for each, \ie we have $K{\sxtimes}C$-many centers for a $C$-class problem. Next, we refer to nearest-neighbor classifiers and obtain an assignment map of dimension $(K{\sxtimes}C)\sxtimes(K{\sxtimes}C)$ as $p_{\mathcal{T}}$ to express their shared and discriminative relationship for all sub-classes, see Alg. 2 and Sec. ? in supplementary material for formal formulation. During training, different to data-driven annotations, we use the same annotation for all the features belonging to the same sub-class. Hence, the transferred knowledge represents the teacher's local decision capacity rather than its feature geometry. 

\input{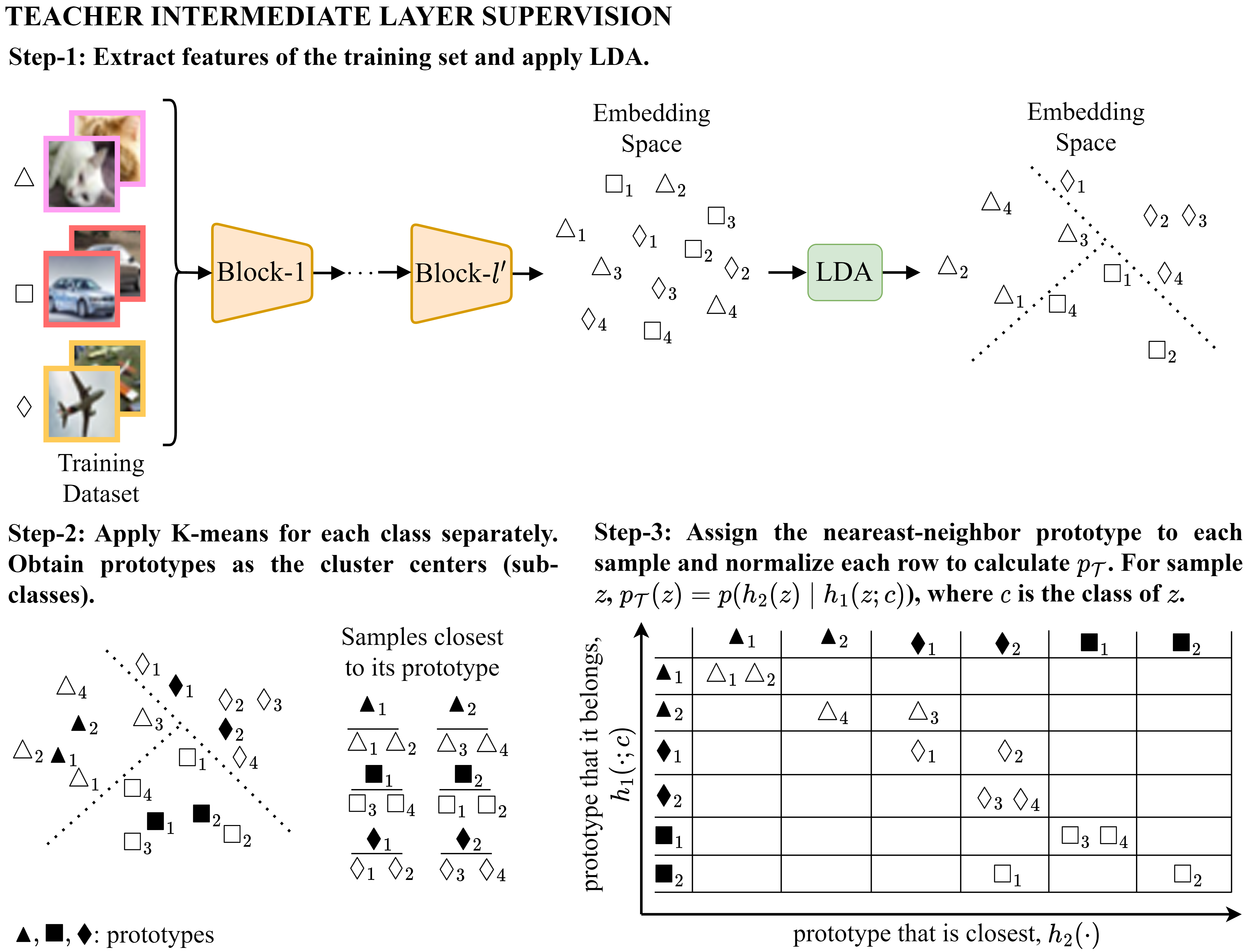}

Surely, we expect some features to be shared among different classes (\eg \emph{tire} for \emph{truck} and \emph{car}) while some are discriminative (\eg \emph{beak} for \emph{bird}). To capture such localized decisions, we apply $K$-means of small $K$ to the features of each class separately and obtain $K$-many sub-class centers for each, \ie we have $K{\sxtimes}C$-many centers for a $C$-class problem. Next, we consider two nearest-neighbor classifiers: $h_1(\cdot; c)$ that labels its input with the ID of the nearest center belonging to class $c{\in}[C]$, and $h_2(\cdot)$ that assigns the ID of the nearest center to its input. Using the dataset, we estimate the label distribution of the features belonging to a particular sub-class. Namely, we estimate the probability that $h_2$ assigns $j$ when $h_1$ assigns $i$ and obtain the distribution $p(h_2(\cdot) \mid h_1(\cdot; c))$, which allows us to see whether the teacher finds it useful to discriminate some sub-classes at layer-$l^\prime$. During training, we pass the input, $(x,y)$, through the teacher to have $z\coloneqq \mathrm{LDA}(f_t^{(l^\prime)}(x;\theta_t))$, and obtain the soft labeling for each pixel-$i$ of the feature map as $p_{\mathcal{T}}(i)= p(h_2(z_i)\mid h_1(z_i, y))$ {(\ie \textit{rows} of the table in \cref{fig:intermediate_kd})}. Different from data-driven annotations, we use the same annotation for all the features belonging to the same sub-class. Hence, the transferred knowledge represents the teacher's local decision capacity rather than its feature geometry.

\section{Experimental Work}
%We start our empirical study with evaluations on KD classification benchmarks to examine the effectiveness of our KD layer for various models. We extend our study further to validate the role of our KD layer in learning. 

%\subsection{Setup}
\label{sec:setup}

% Reducing the confounding of the factors other than our method, we adopt the framework\footnote{We defer the details to supplementary material.} implemented by \cite{jain2019quest} in PyTorch \cite{NEURIPS2019_9015} to make a fair and unbiased evaluation of our method as well as comparisons with the other invented methods. We evaluate our method on CIFAR-100 \cite{krizhevsky2009learning}, Tiny-ImageNet \cite{Le2015TinyIV} and ImageNet \cite{imagenet} with various architectures including ResNet (RN) \cite{HeResnetv1_2015,he2016identity}, Wide ResNet (WRN) \cite{zagoruyko2016wide}, MobileNet (MNV2) \cite{MobileNetV2} and ShuffleNet (SNV1/V2) \cite{ShufflenetV1,ShufflenetV2}. We attribute our methods as letKD-1 and letKD-2, where letKD-1 represents the single penultimate layer KD while letKD-2 denotes inclusion of intermediate layer KD. For the intermediate layer, we place our KD layer after the first residual block of the last stage for all networks, and use the output of the first block at the last stage of the teacher. 

Reducing the confounding of the factors other than our method, we adopted the framework implemented by \cite{jain2019quest} in PyTorch \cite{NEURIPS2019_9015} to make a fair and unbiased evaluation of our method as well as comparisons with the other invented methods. We evaluated our method on CIFAR-100 \cite{krizhevsky2009learning}, Tiny-ImageNet \cite{Le2015TinyIV} and ImageNet \cite{imagenet} with various architectures including ResNet (RN) \cite{HeResnetv1_2015,he2016identity}, Wide ResNet (WRN) \cite{zagoruyko2016wide}, MobileNet (MNV2) \cite{MobileNetV2} and ShuffleNet (SNV1/V2) \cite{ShufflenetV1,ShufflenetV2}. We attribute our methods as letKD-1 and letKD-2, where letKD-1 represents the single penultimate layer KD while letKD-2 denotes the inclusion of intermediate layer KD. {As for selecting the location of the layers in our methods, we deliberately used the penultimate layer, a common
choice in KD methods, as it carries the most discriminative
and higher-level information. For the intermediate layer, we relied on TDD’s findings
\cite{Song_2021_CVPR} indicating class-agnostic characteristics of lower
layer teacher’s features and thus, omitted the lower layers and decided to place our KD layer after the first residual block of the last stage for all networks, and used the output of the first block at the last stage of the teacher. Additionally, this choice ensures enough receptive field for
the validity of our template matching formulation defined in \eqref{eq:argmax}.} We defer further empirical details including hyperparameter selection and its analysis to \refsuppl.

\subsection{Results}
% We provide the results in \cref{tab:cifar100,tab:imagenet,tab:tinyimagenet} for the evaluations on CIFAR-100, ImageNet and Tiny-ImageNet, respectively. We compare our method against \cite{Hinton2015}, FitNet \cite{FitNet}, AT \cite{zagoruyko2017paying}, AB \cite{heo2019AB}, FSP \cite{yim2017FSP}, SP \cite{Tung_2019_ICCV}, VID \cite{Ahn_2019_CVPR}, CRD \cite{tian2019crd}, DKD \cite{Zhao_2022_CVPR}, SimKD \cite{Chen_2022_CVPR}, TDD \cite{Song_2021_CVPR} and QUEST \cite{jain2019quest}. 

We provide the results in \cref{tab:cifar100,tab:imagenet} for the evaluations on CIFAR-100, and ImageNet, respectively. We compare our method against KD \cite{Hinton2015}, FitNet \cite{FitNet}, DKD \cite{Zhao_2022_CVPR}, SimKD \cite{Chen_2022_CVPR}, TDD \cite{Song_2021_CVPR} and QUEST \cite{jain2019quest}. We defer the results for Tiny-ImageNet and the extended versions of \cref{tab:cifar100,tab:imagenet} with other KD alternatives to the supplementary material \refextendedemp. 

% \textbf{CIFAR-100 (\cref{tab:cifar100}).} Our method letKD-2 outperforms all the other methods on all different teacher-student combinations with homogeneous and heterogeneous architecture settings, except for RN32x4-RN8x4 where we are slightly behind SimKD \cite{Chen_2022_CVPR}. With that being said, SimKD includes additional $1{\sxtimes}1$-$3{\sxtimes}3$-$1{\sxtimes}1$ convolution block before penultimate layer, which captures further local relations especially in shallow architectures. When we compare our method with their $1{\sxtimes}1$-$1{\sxtimes}1$ version, we outperform them by $\approx 1.5\%$ points margin. Moreover, simKD is inferior in all the other teacher-student pairs even with their largest model, yet another supporting evidence for the effectiveness of our KD layer. 

\textbf{CIFAR-100 (\cref{tab:cifar100}).} Our method letKD-2 outperforms all the other methods on all different teacher-student combinations with homogeneous and heterogeneous architecture settings with letKD-1 being the second best, except for RN32x4-RN8x4 where we are slightly behind SimKD \cite{Chen_2022_CVPR}. With that being said, SimKD includes an additional $1{\sxtimes}1$-$3{\sxtimes}3$-$1{\sxtimes}1$ convolution block before the penultimate layer, which captures further local relations. When we compare our method with their $1{\sxtimes}1$-$1{\sxtimes}1$ version, we outperform them by $\approx 1.3\%$ points margin. Moreover, SimKD is inferior in all the other teacher-student pairs even with their largest block, yet another supporting evidence for the effectiveness of our KD layer.

\input{figures/table_CIFAR100.tex}
\input{figures/table_ImageNet}

\textbf{ImageNet(\cref{tab:imagenet}).} We verify the scalability of our methods by performing experiments on large-scale datasets including ImageNet. Overall, our method letKD-2 outperforms all the other methods on all the different teacher-student combinations with homogeneous and heterogeneous architecture settings. 

% Finally, we also observe that letKD-1 also shows a relatively good performance and can be considered the second best with respect to other methods (especially regarding the QUEST method itself) for all datasets except for SimKD in CIFAR-100, which we disclose the factors behind. All of these results rigorously show that forcing local regions to be semantically represented from the combination of the soft labels of the teacher would yield better predictions in both training and inference through the proposed KD layer.

\subsection{Ablations and Behavior Analysis}
\label{sec:ablation}
\textbf{With and without $K$-means.} To validate our analysis in \cref{sec:pen_labelling} that repurposes the $3{\sxtimes}3$ kernels of a $3{\sxtimes}3$-BN-ReLU-$1{\sxtimes}1$  block as the cluster centers for the semantic entities, we evaluate our method on CIFAR-100 dataset with soft assignments $p_\mathcal{T}$ computed using the $3{\sxtimes}3$ kernels as explained in \cref{sec:pen_labelling} and the clusters obtained through $K$-means. We denote the experiments with (without) our KD layer as letKD-1 (QUEST) for both cases. 

\input{figures/table_3x3_vs_QUEST}

The results provided in  \cref{tab:3x3_vs_quest} show that the exploitation of $3{\sxtimes}3$ kernels can replace $K$-means based supervision. We observe competitive and even better performances with $3{\sxtimes}3$ kernels as the cluster centers. Such results validate our claims suggesting that $3{\sxtimes}3$ kernels inherently learn templates (\ie cluster centers) that are capturing some semantic entities owing to the template matching paradigm introduced in \cref{sec:kdlayer_formulation}.

\input{figures/table_quest_multi_layer.tex}

\textbf{Let the student decide.} We argue that using the soft labels obtained by $K$-means clustering of the teacher's features in the intermediate layers eventually results in pushing the student to imitate the teacher's geometry. As argued previously \cite{Song_2021_CVPR}, the student can fail to effectively exploit that geometry due to capacity differences. Nonetheless, our KD layer is equipped with a mechanism to discard the information that may have a detrimental effect on the performance. To validate, we evaluate the multi-layer version of our method with $K$-means supervision in both layers, denoted as letKD-1 (2). We also evaluate QUEST under the same setting, denoted as QUEST (2). In the intermediate layer, we apply a temperature to the logits before obtaining $p_\mathcal{T}$ for both cases. The results in \cref{tab:quest_multi} on CIFAR-100 with RN56-RN20 show that QUEST with intermediate KD degrades the performance while our layer successfully shapes the knowledge to fit its representation capability.

% To implement their best version, we apply the same temperature to the logits before obtaining $p_\mathcal{T}$ for both cases. 

\textbf{Effect of the KD layer.} Towards the understanding of the impact of our KD layer for intermediate layer (\ie lower level) supervision, we measured the classification capacity of the student trained with our methods. We achieved this by fitting a linear classifier to the global features of the trained student at that layer. Owing to this study with the results tabulated in the supplementary material \refsupptabfour, we showed that our KD layer can improve the student's decision capacity significantly. To further validate our KD layer's ability to shape the intermediate features of the student by exploiting the teacher's knowledge, we analyzed the effect of how enhancing the student's features with the weighted combinations of the learned semantic vectors improves the overall performance. Namely, we set $\alpha=0$ in \eqref{eq:kd_layer} to lift the knowledge-based feature transform and compare its performance with $\alpha=1$. The results presented in the supplementary material \refsupptabfive show consistent improvement of the inclusion of our KD layer. Finally, we addressed whether the performance increase is coming from the method or the capacity increase introduced by our KD layer. As summarized in the supplementary material \refsupptabsix, we compared the performance of the three methods as FitNet, FitNet equipped with our KD layer at the penultimate layer with and without supervision to show that even though the capacity of the student is increased due to our KD layer, the major contribution for the performance occurs by combining it with our supervision.

\section{Conclusion}
We bring a different perspective to KD formulation in terms of a portable residual layer that improves KD by explicitly embedding the teacher's knowledge in feature transform. This way, we enable the student to discard nuisance information and feed forward transferred knowledge deeper for improved inference. To facilitate knowledge transfer in the intermediate layers, we also propose a novel form of supervision based on teacher's decisions. With extensive empirical studies, we validated the effectiveness of the proposed KD layer in various KD benchmarks.

% References
% -----------
\bibliography{library}

% Supplementary
% -------------
\ifx \printOption \printArxiv
\input{supplementary/supplementary_append}

\fi

\end{document}

%% file: packages.tex
\usepackage{import}
\usepackage[utf8]{inputenc} % allow utf-8 input
\usepackage[T1]{fontenc}    % use 8-bit T1 fonts
\usepackage{lmodern}

\DeclareGraphicsExtensions{.pdf,.jpeg,.png} % and their extensions so you won't have to specify these with every instance of \includegraphics
\usepackage{wrapfig}
\usepackage{tabularx}
\usepackage{booktabs, multicol, multirow}	% professional-quality tables
\usepackage{array}
\usepackage{xcolor}
\usepackage{algorithmicx}
\usepackage[noend]{algpseudocode}
\usepackage{eqparbox}
\usepackage{algorithm}
\usepackage{amsmath}
\usepackage{mathtools}
\usepackage{commath}
\usepackage{mathrsfs}
\usepackage{amssymb}
\usepackage{amsfonts}       % blackboard math symbols
\usepackage{bbm}
\usepackage{bm}
\usepackage{fancyhdr}
\usepackage{cancel}
\usepackage{nicefrac}       % compact symbols for 1/2, etc.
\usepackage{microtype}      % microtypography
\usepackage{mdwmath}
\usepackage{mdwtab}
\usepackage{wasysym} % for box symbol

\usepackage[capitalize]{cleveref}
\crefname{section}{\S}{\S\S}
\Crefname{section}{Section}{Sections}
\Crefname{table}{Table}{Tables}
\crefname{table}{Tab.}{Tabs.}

\usepackage{url}            % simple URL typesetting

%% file: definitions.tex
\def\codeurl{https://github.com/adagorgun/letKD-framework} % code url

\def\printArxiv{arxiv}
\DeclareMathOperator*{\argmax}{arg\,max}

\newcommand{\R}{\ensuremath{\mathbb{R}}}

\DeclareMathSymbol{\shortminus}{\mathbin}{AMSa}{"39}

\newcommand{\sxtimes}{\mathsf{x}}%\mskip1mu
\newcommand{\T}{\ensuremath{^\intercal}}
\newcommand{\ie}{\textit{i}.\textit{e}., }
\newcommand{\eg}{\textit{e}.\textit{g}., }

\numberwithin{equation}{section}

\makeatletter
\def\BState{\State\hskip-\ALG@thistlm}
\makeatother

%% file: frontmatter.tex
\title{Knowledge Distillation Layer that Lets the Student Decide}
\def\etal{\emph{et al}\bmvaOneDot}

\runninghead{G\"{o}rg\"{u}n \MakeLowercase{\etal}}{KD Layer that Lets the Student Decide}
\input{authors}

%% file: authors.tex
\def\ada{Ada~G\"{o}rg\"{u}n}
\def\yeti{Yeti~Z.~G\"{u}rb\"{u}z}

\def\aydin{A.~Ayd{\i}n~Alatan}
\def\inst1{Dept. of Electrical and Electronics Eng., Middle East Technical University, Ankara, Turkey}

\addauthor{\ada\textsuperscript{*}}{ada.gorgun@metu.edu.tr}{1}
\addauthor{\yeti\textsuperscript{*}}{yeti@metu.edu.tr}{1}
\addauthor{\aydin}{alatan@metu.edu.tr}{1}

\addinstitution{
 Dept. of Electrical and Electronics Eng. \& Center for Image Analysis (OGAM)\\
 Middle East Technical University\\
 Ankara, Turkey
}

%% file: abstract.tex
\begin{abstract}
Typical technique in knowledge distillation (KD) is regularizing the learning of a limited capacity model (\emph{student}) by pushing its responses to match a powerful model's (\emph{teacher}). Albeit useful especially in the penultimate layer and beyond, its action on student's feature transform is rather implicit, limiting its practice in the intermediate layers. To explicitly embed the teacher's knowledge in feature transform, we propose a learnable KD layer for the student which improves KD with two distinct abilities: \emph{i)} learning how to leverage the teacher's knowledge, enabling to discard nuisance information, and \emph{ii)} feeding forward the transferred knowledge deeper. Thus, the student enjoys the teacher's knowledge during the inference besides training. Formally, we repurpose $1{\sxtimes}1$-BN-ReLU-$1{\sxtimes}1$ convolution block to assign a semantic vector to each local region according to the template (supervised by the teacher) that the corresponding region of the student matches. To facilitate template learning in the intermediate layers, we propose a novel form of supervision based on the teacher's decisions. Through rigorous experimentation, we demonstrate the effectiveness of our approach on 3 popular classification benchmarks. Code is available at: \href{\codeurl}{letKD Framework}
\end{abstract}

%\begin{keyword}
%ResNet \sep arg-max \sep feature embedding \sep template matching \sep classification
%\end{keyword}

%\begin{keywords}
%ResNet, arg-max, embedding
%\end{keywords}

%% file: arxiv_addon.tex
\ifx \printOption \printArxiv
% first page header
\fancypagestyle{firststyle}
{
   \fancyhead{}
   \lhead{Accepted as a conference paper at BMVC 2023}
}
\thispagestyle{firststyle}
% alias for references to supp
\def\refappendix{\hyperref[sec:bnargmax]{appendix}}
\def\refalgos{with \Cref{algo:algo_penultimate,algo:algo_student}}
\def\refinteralgo{with \Cref{algo:algo_intermediate} }
\def\refsuppl{the supplementary material}
\def\refextendedemp{\cref{sec:sup_extend_eval}}

\def\refsupptabfour{\cref{tab:subclass_value_offline}}
\def\refsupptabfive{\cref{tab:value_effect} }
\def\refsupptabsix{\cref{tab:fitnet}}

\else
\def\refappendix{the supplementary material \citep[Appendix]{gorgun_appendix}}
\def\refalgos{\citep[\S~3]{gorgun_appendix} with Algorithms~1 and 3}
\def\refinteralgo{\citep[\S~3]{gorgun_appendix} with Algorithm~2 }
\def\refsuppl{the supplementary material \citep{gorgun_appendix}}
\def\refextendedemp{\citep[\S~1.1]{gorgun_appendix}}
\def\refsupptabfour{\citep[Tab.~4]{gorgun_appendix}}
\def\refsupptabfive{\citep[Tab.~5]{gorgun_appendix} }
\def\refsupptabsix{\citep[Tab.~6]{gorgun_appendix}}

\fi

%% file: figures/fig1.tex
% \begin{figure}[t]
%   \centering
%   \centerline{\includegraphics[width=.5\linewidth]{figures/fig1_total.png}}
%   \caption{Illustration of our method-letKD. Typical KD methods use regularization while pushing the student to imitate the feature geometry of the teacher. We learn the semantic entities that the teacher finds useful and exploit them in feature transform, enabling us to feed forward the knowledge.}
% 	\label{fig:fig1}
%   \end{figure}

\begin{wrapfigure}[20]{r}[-3pt]{0.46\linewidth} % 15
 \vspace{-1.4\intextsep} %\vspace{-2.5\intextsep}
  %\centering
  \centerline{\includegraphics[width=\linewidth]{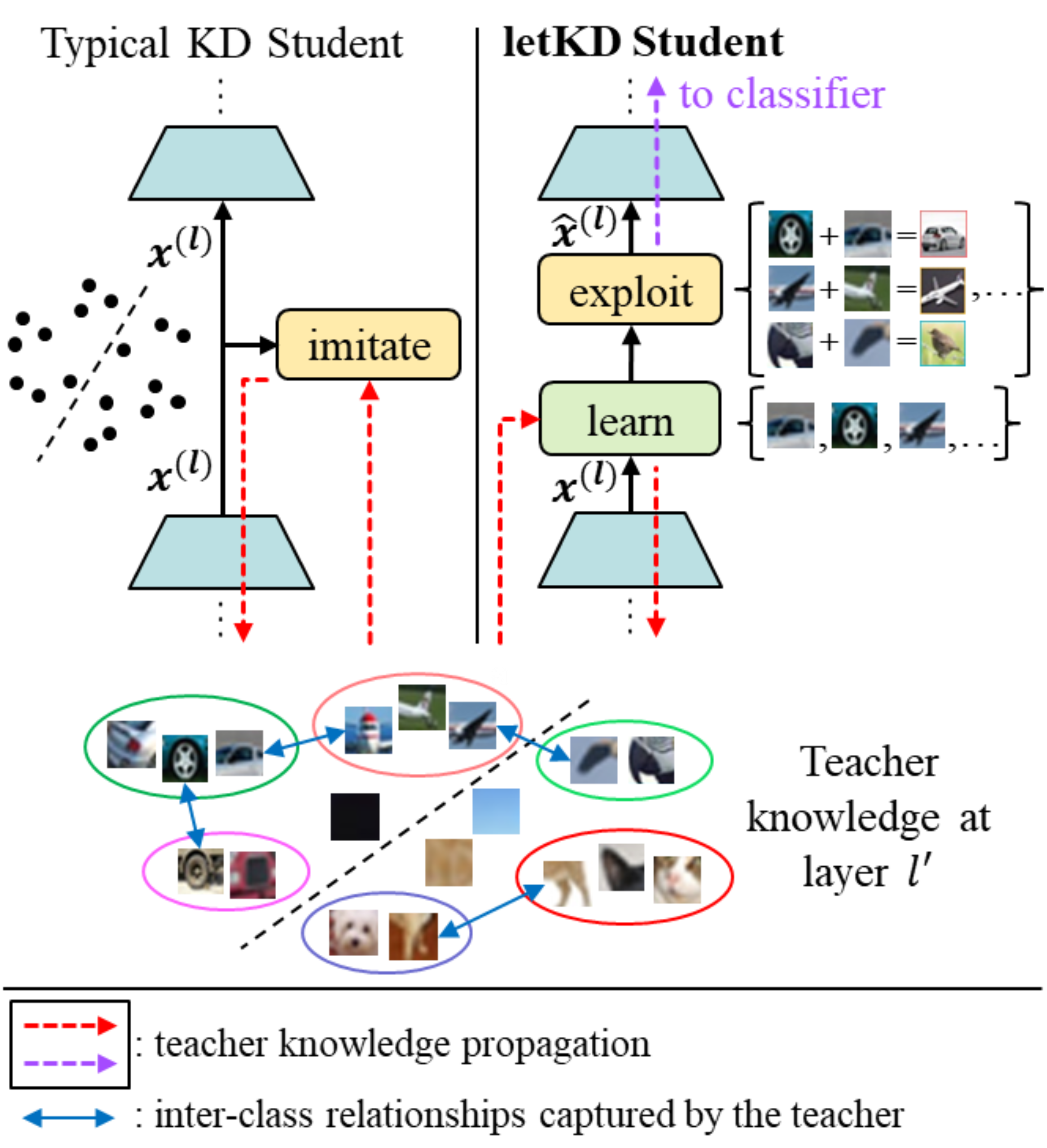}}
  \caption{The differences between training the student with typical KD methods and with letKD.}
	\label{fig:fig1}
\end{wrapfigure}

% Illustration of our method-letKD. Typical KD methods use regularization while pushing the student to imitate the feature geometry of the teacher. We learn the semantic entities that the teacher finds useful and exploit them in feature transform, enabling us to feed forward the knowledge.

%% file: figures/fig_method.tex
% \begin{figure}[t]
%   \centering
%   \centerline{\includegraphics[width=.7\linewidth]{figures/fig_method.png}}
%   \caption{Proposed learnable KD layer}
% 	\label{fig:kdlayer}
%   \end{figure}

\begin{wrapfigure}[10]{r}[-5pt]{0.5\linewidth} % 15
% \vspace{-4.5\intextsep} 
\vspace{-2.\intextsep}
  %\centering
  \centerline{\includegraphics[width=\linewidth]{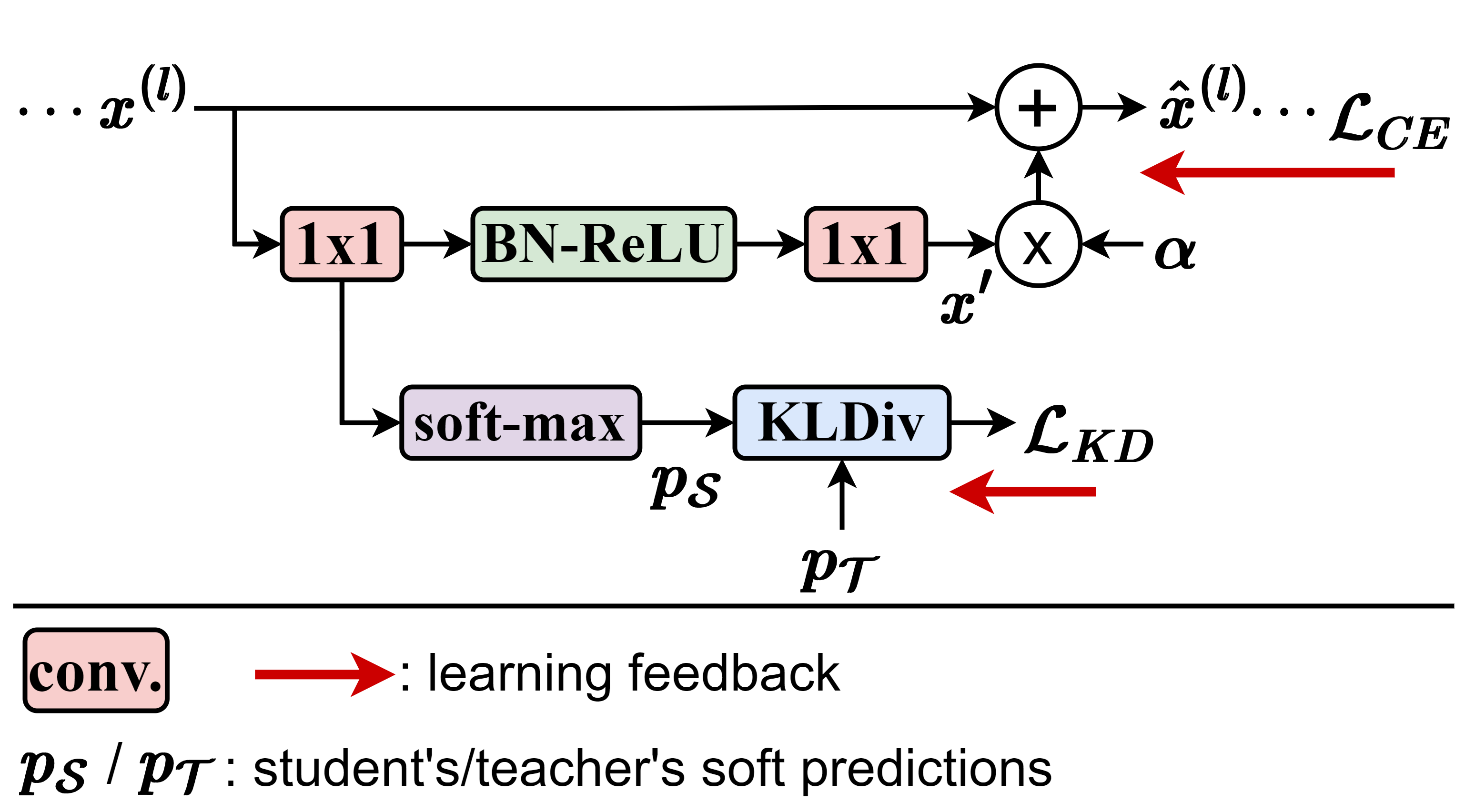}}
  \caption{Proposed KD layer.}
	\label{fig:kdlayer}
\end{wrapfigure}

%Proposed learnable KD layer with $p_{\mathcal{S}}$ and $p_{\mathcal{T}}$ are the soft-predictions of student  and teacher.

% \begin{wrapfigure}[6]{r}[-2pt]{0.43\linewidth} % 15
% \vspace{-4.5\intextsep} 
% %\vspace{-2.5\intextsep}
%   %\centering
%   \centerline{\includegraphics[width=\linewidth]{figures/fig_method_new.png}}
%   \caption{Proposed KD layer.}
% 	\label{fig:kdlayer}
% \end{wrapfigure}

%% file: figures/intermediate_kd.tex
%\vspace{-1.5mm}
\begin{figure}[!h]
  \centering
  \centerline{\includegraphics[width=.81\linewidth]{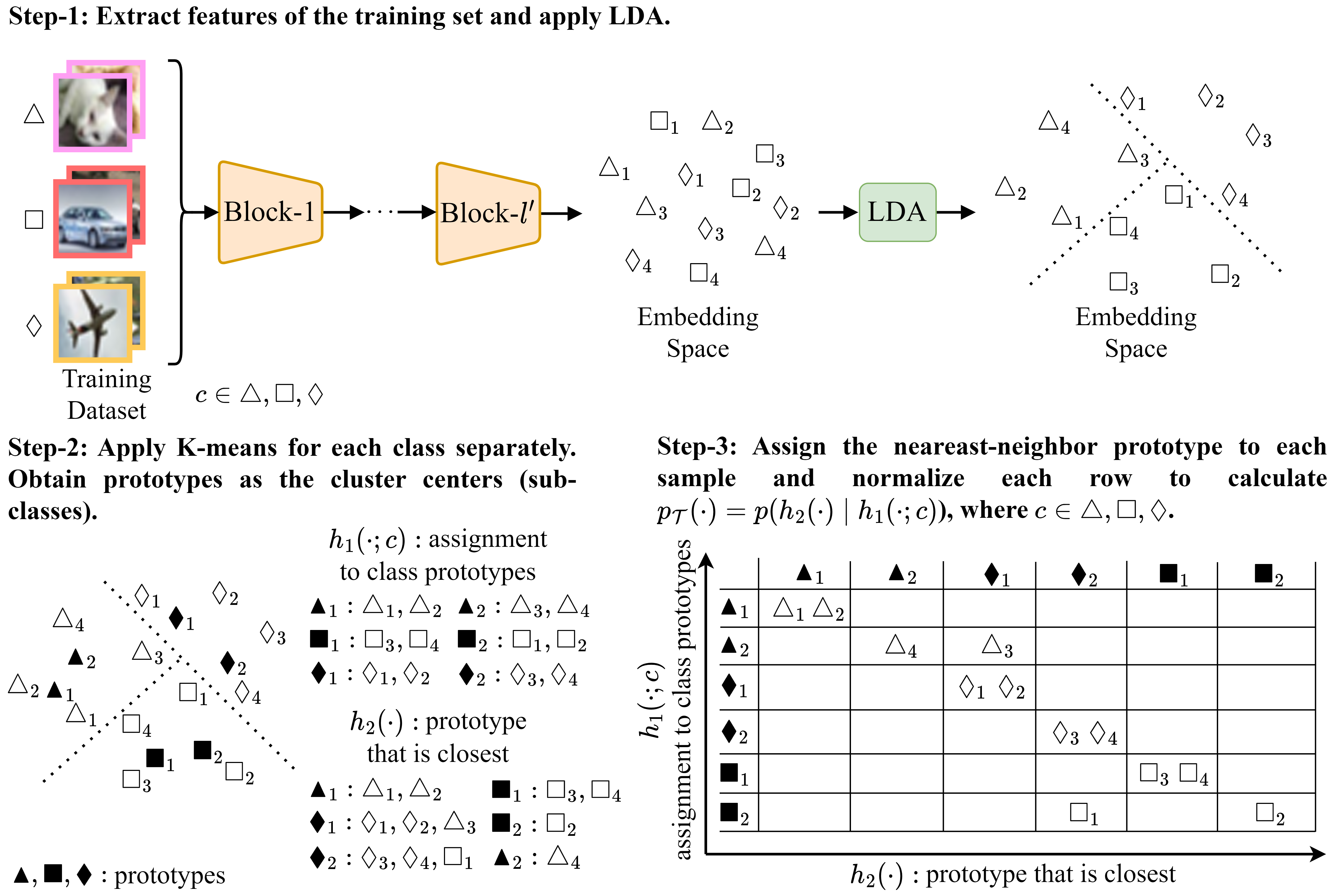}}
  \caption{Visualization of the teacher's intermediate layer supervision.}
	\label{fig:intermediate_kd}
\end{figure}

%% file: figures/table_CIFAR100.tex
% \usepackage{graphicx}

\begin{table*}[ht]%[13]{r}[-5pt]{0.6\linewidth}
    %\vspace{-.9\intextsep}
	\centering
	\caption{Top-1 acc. averaged over 5 trials on CIFAR100. \textbf{Bold}: best in its category.}
	\label{tab:cifar100}
	\resizebox{\linewidth}{!}{%
    	\begin{tabular}{cccccccccccc}
    		\toprule 
    		Archs. $\rightarrow$ & \multicolumn{6}{c}{\textbf{Homogeneous}} & & \multicolumn{4}{c}{\textbf{Heterogeneous}} \\ \cmidrule{2-7} \cmidrule{9-12} 
    		Teacher & WRN-40-2 & WRN-40-2 & RN56 & RN110 & RN110 & RN32x4 & & WRN-40-2 & RN32x4 & RN32x4 & RN50\\  
    		Student & WRN-16-2 & WRN-40-1 & RN20 & RN20  & RN32  & RN8x4 & & SNV1 & SNV1 & SNV2 & MNV2\\ \midrule
    		\multirow{2}{*}{Methods $\downarrow$} & 75.61 & 75.61 & 72.34 & 74.31 & 74.31 & 79.42 & & 75.61 & 79.42 & 79.42 & 79.34 \\
    		 & 73.26 & 71.98 & 69.06 & 69.06 & 71.14 & 72.50 & & 70.50 & 70.50 & 71.82 & 64.60 \\ \midrule
            KD & 74.92 & 73.54 & 70.66 & 70.67 & 73.08 & 73.33 & & 74.83 & 74.07 & 74.45 & 67.35\\
            FitNet & 73.58 & 72.24 & 69.21 & 68.99 & 71.06 & 73.50 & & 73.73 & 73.59 & 73.54 & 63.16 \\
            DKD & 76.24 & 74.81 & 71.97 & - & 74.11 & 76.32 & & 76.70 & 76.45 & 77.07 & 70.35 \\
            SimKD & 76.06 & 74.92 & 68.95 & 69.35 & 72.15 & \textbf{78.08} & & 76.95 & 77.18 & 77.78 & 68.91 \\
            TDD & 75.01 & 74.04 & 71.53 & - & - & - & & 75.60 & - & - & 68.37 \\
            QUEST & 76.10 & 74.58 & 71.84 & 71.89 & 74.08 & 75.88 & & 76.75 & 76.28 & 77.09 & 69.81 \\
            \midrule
            \textbf{letKD-1} & 
            $\underset{\scriptsize{\mp0.15}}{76.29}$ & $\underset{\scriptsize{\mp0.09}}{75.01}$ & $\underset{\scriptsize{\mp0.24}}{72.44}$ & $\underset{\scriptsize{\mp0.31}}{72.68}$ & $\underset{\scriptsize{\mp0.14}}{74.40}$ & $\underset{\scriptsize{\mp0.06}}{76.70}$ & & $\underset{\scriptsize{\mp0.16}}{76.93}$ & $\underset{\scriptsize{\mp0.24}}{76.65}$ & $\underset{\scriptsize{\mp0.17}}{77.75}$ &
            $\underset{\scriptsize{\mp0.18}}{69.97}$ \\
            \textbf{letKD-2} & 
    $\mathbf{\underset{\scriptsize{\mp0.22}}{76.56}}$ & $\mathbf{\underset{\scriptsize{\mp0.13}}{75.19}}$ & $\mathbf{\underset{\scriptsize{\mp0.16}}{73.27}}$ & $\mathbf{\underset{\scriptsize{\mp0.14}}{73.38}}$ & $\mathbf{\underset{\scriptsize{\mp0.20}}{74.62}}$ & $\mathbf{\underset{\scriptsize{\mp0.18}}{77.09}}$ & & $\mathbf{\underset{\scriptsize{\mp0.12}}{77.08}}$ & $\mathbf{\underset{\scriptsize{\mp0.12}}{77.30}}$ & $\mathbf{\underset{\scriptsize{\mp0.06}}{77.95}}$ &
    $\mathbf{\underset{\scriptsize{\mp0.23}}{70.39}}$ \\
    		\bottomrule
    	\end{tabular}%
    }
\end{table*}

%% file: figures/table_ImageNet.tex
% \usepackage{graphicx}

\begin{table*}[ht]%[13]{r}[-5pt]{0.6\linewidth}
    \vspace{-.9\intextsep}
	\centering
	\caption{Top-1 and top-5 acc. on ImageNet. Teacher-Student \textbf{(a)}: RN34-RN18 \textbf{(b)}: RN50-MNV2. \textbf{Bold}: best in its category.}
	\label{tab:imagenet}
        \resizebox{\textwidth}{!}{
	\begin{tabular}{c|cccccccc}
		\toprule 
		Setting & & Teacher & Student & KD & DKD & QUEST & \textbf{letKD-1} & \textbf{letKD-2} \\ \midrule   
		\multirow{2}{*}{(a)} & Top-1 & 73.31 & 69.75 & 70.66 & 71.70 & 71.67 & 72.33 & \textbf{72.38} \\
		& Top-5 & 91.42 & 89.07 & 89.88 & 90.41 & 90.67 & 91.06 & \textbf{91.15}\\ \midrule
		\multirow{2}{*}{(b)} & Top-1 & 76.13 & 68.87 & 68.58 & 72.05 & 72.54 & 73.78 & \textbf{73.98}\\
		& Top-5 & 92.86 & 88.76 & 88.98 & 91.05 & 91.13 & 91.81 & \textbf{92.00}\\
		\bottomrule
	\end{tabular}%
        }
\end{table*}

%% file: figures/table_3x3_vs_QUEST.tex
% \usepackage{graphicx}

% \begin{table*}[ht]%[13]{r}[-5pt]{0.6\linewidth}
% \begin{table}[H]
% 	\centering
% 	\caption{}
% 	\label{tab:3x3_vs_quest}
% 	\resizebox{\linewidth}{!}{%
%     	\begin{tabular}{c|c|c|c|c|cc}
%     		\hline 
%     		  Words $\rightarrow$ & \multicolumn{2}{c|}{3x3 Kernels} & \multicolumn{2}{c|}{K-Means} & \multirow{2}{*}{Teacher} & \multirow{2}{*}{Student} \\ \cline{2-5}
%              Archs. $\downarrow$ & \multicolumn{1}{c|}{\textbf{w/o}} & \multicolumn{1}{c|}{\textbf{w}} & \multicolumn{1}{c|}{\textbf{w/o}} & \multicolumn{1}{c|}{\textbf{w}} & & \\ \hline
%              RN56-RN20 & 71.92 & 72.11 & 71.84 & 72.44 & 72.34 & 69.06 \\ \hline
%              RN110-RN32 & 74.31 & 74.44 & 74.08 & 74.40 & 74.31 & 71.14 \\ \hline
%              RN83-RN29 & 72.41 & 72.61 & 72.48 & 73.33 & 73.84 & 70.53 \\ \hline		
%     	\end{tabular}%
%     }
% \end{table}
% \end{table*}

\begin{wraptable}[6]{r}[-3pt]{0.6\linewidth}
    \vspace{-1.25\intextsep}	
    \centering
	\caption{Average top-1 accuracies on CIFAR-100 over 5 trials to validate the analysis on $3{\sxtimes}3$ kernels.}
	\label{tab:3x3_vs_quest}
	\resizebox{\linewidth}{!}{%
    	\begin{tabular}{cccccccc}
    		\toprule 
    		 Clusters $\rightarrow$ & \multicolumn{2}{c}{3x3 Kernels} & & \multicolumn{2}{c}{K-Means} & \multirow{2}{*}{Teacher} & \multirow{2}{*}{Student} \\ \cmidrule{2-3} \cmidrule{5-6}
             Archs. $\downarrow$ & QUEST & letKD-1 & & QUEST & letKD-1 & & \\ \midrule
             RN56-RN20 & 71.92 & 72.11 & & 71.84 & 72.44 & 72.34 & 69.06 \\ 
             RN110-RN32 & 74.31 & 74.44 & & 74.08 & 74.40 & 74.31 & 71.14 \\ 
             RN83-RN29 & 72.41 & 72.61 & & 72.48 & 73.33 & 73.84 & 70.53 \\ \bottomrule		
    	\end{tabular}%
    }
\end{wraptable}

%% file: figures/table_quest_multi_layer.tex
% \usepackage{graphicx}

% \begin{table*}[ht]%[13]{r}[-5pt]{0.6\linewidth}
% \begin{table}[H]
\begin{wraptable}[10]{r}[-3pt]{0.4\linewidth}
    \vspace{-.1\intextsep}
	\centering
	\caption{Effect of multi-layer distillation with QUEST and letKD-1.}
	\label{tab:quest_multi}
	\begin{tabular}{lc}
		\toprule 
		Method & Top-1 Acc.\\ \midrule
        QUEST & 71.84 \\
        QUEST (2) & 71.79 \\
        letKD-1 & 72.44 \\
        letKD-1 (2) & 72.73 \\
		\bottomrule
	\end{tabular}%
% \end{table*}
% \end{table}
\end{wraptable}

%% file: supplementary/supplementary_append.tex
% Title
% ------
\setcounter{section}{0}
\section*{\Large  \textbf{Supplementary Material for {\emph{"Knowledge Distillation Layer that Lets the Student Decide"}}}}

% Body
% -----
\input{supplementary/body}

\input{supplementary/appendix}

%% file: supplementary/body.tex
% Paper Part
% ----------

\section{Extended Empirical Study}
\label{sec:sup_extende_emp}

% \yetinote{Buraya kisa da olsa bir seyler yazmaliyiz direkt 1.1 gelmesin.}
% \yetinote{bu ek deneyler basa gelmeli bence.}
In this section, we provide comparisons of our method to additional KD methods, along with the evaluation of our method on Tiny-ImageNet. We also provide more ablations to offer deeper insights into the efficacy of our method.

\subsection{Results on More Datasets and Methods}
\label{sec:sup_extend_eval}
% \yetinote{Burada table basligin hemen altina gelmesin sayfa altinda olsun.}
In this section, we provide the extended results in \cref{tab:cifar100_sup,tab:imagenet_sup,tab:tinyimagenet} for the evaluations on CIFAR-100 \cite{krizhevsky2009learning}, ImageNet \cite{imagenet} and Tiny-ImageNet \cite{Le2015TinyIV}, respectively. We compare our method against KD \cite{Hinton2015}, FitNet \cite{FitNet}, AT \cite{zagoruyko2017paying}, AB \cite{heo2019AB}, FSP \cite{yim2017FSP}, SP \cite{Tung_2019_ICCV}, VID \cite{Ahn_2019_CVPR}, CRD \cite{tian2019crd}, DKD \cite{Zhao_2022_CVPR}, SimKD \cite{Chen_2022_CVPR}, TDD \cite{Song_2021_CVPR} and QUEST \cite{jain2019quest}. Overall, the results demonstrate the effectiveness of our methods letKD-2 and letKD-1 for all the datasets by being the first and the second best compared to other KD alternatives except with the SimKD method in RN32x4-RN8x4 on CIFAR-100 for which the relevant discussion is provided in the main paper.

\input{supplementary/figures/table_CIFAR100}
\vspace{-1.4\intextsep}
\input{supplementary/figures/table_ImageNet}
\input{supplementary/figures/table_TinyImageNet}

\subsection{More Ablations}
\label{sec:sup_more_ablation}

\input{supplementary/figures/fig_subclass_number}
\textbf{Hyperparameters.} Our KD layer introduces 4 additional hyperparameters which are $\lbrace \alpha_{inter}, \alpha_{penult} \rbrace$ from Fig. 2 in the main paper, $K_{penult}$ the number of cluster centers used in the penultimate layer, and $K_{inter}$ the number of sub-classes for each class used in the intermediate layer. We use $K_{penult}=4096$ to be directly comparable with \cite{jain2019quest}. For $K_{inter}$, through our analysis on CIFAR-100 with RN56-RN20, plotted in \cref{fig:subclass_number}, we observe a relatively stable performance for $K_{inter}>8$. Hence, we set $K_{inter}=8$ for the rest of the experiments. Finally, for the selection of $\lbrace \alpha_{inter}, \alpha_{penult} \rbrace$, since we use normalized convolution with a learnable scale for the $1\sxtimes1$ to jointly learn it (\cref{sec:experiment_details}), we set $(\alpha_{inter}, \alpha_{penult})=(1,1)$ in all models except for RN50-MNV2 in CIFAR-100 and Tiny-ImageNet. Essentially, in both letKD-1 and letKD-2 experiments, \emph{(i)} for CIFAR-100, we arrange $\alpha_{penult}=0.1$ and $\alpha_{inter}=0.2$, \emph{(ii)} for Tiny-ImageNet, we arrange $\alpha_{penult}=0.5$ and $\alpha_{inter}=1$.

\input{supplementary/figures/table_subclass_value_offline_effect}

\noindent
\textbf{Feature geometry.} Towards the understanding of the impact of the proposed KD layer for intermediate layer (\ie lower level) supervision, we measure the classification capacity of the student trained with our methods in \cref{tab:subclass_value_offline} with $\alpha_{inter}$ representing the inclusion of our layer. In this table, $x$ and $\hat{x}$ represent the input and the output of the KD layer as in Fig. 2 in the main paper. To obtain the classification scores, we perform global average pooling to the extracted intermediate features of the trained student and fit a linear classifier using LDA. This analysis is required to see whether the student is able to exploit the teacher's supervision according to its own discrimination capacity. As seen from \cref{tab:subclass_value_offline}, the student attains 52.71 \% accuracy when it is trying to imitate the teacher ($\alpha_{inter}=0$). When our layer is included ($\alpha_{inter}=1$), even though the input features $x$ perform 1\% poorer compared to $\alpha_{inter}=0$, the output features $\hat{x}$ strongly show the advantage of letting the student freely exploit the teacher rather than directly forcing to imitate it.

\input{supplementary/figures/table_value_effect}

\noindent
\textbf{Effect of the KD layer.}
To further validate exploiting the teacher's knowledge to shape the intermediate features, we analyze the effect of enhancing the student's features with the weighted combinations of the learned semantic vectors. Namely, we set $\alpha=0$ in Fig. 2 in the main paper to lift the knowledge-based feature transform. With the RN56-RN20 pair, we evaluate all possible settings in CIFAR-100 and provide the results averaged over 5 trials in \cref{tab:value_effect}. We observe that $\alpha=1$ consistently improves the performance whereas $\alpha=0$ consistently degrades, especially in the intermediate layer. Based on these results, we can conclude that the student can effectively decide to exploit the semantically meaningful information coming from the teacher to shape the embedding space rather than directly imitating the feature space. The addition of our layer substantially improves the performance, especially when multi-layer supervision is involved. {Nevertheless, it is important to quantitatively show the computational cost of the inclusion of our KD layer. The computation rises merely by adding $1\sxtimes1$-BN-ReLU-$1\sxtimes1$ convolution block, resulting in about 2\% longer sec per image in intermediate layers and 2\%-20\% in the penultimate layer, depending on the number of cluster centers, $K$, ranging from 64 to 4096. Therefore, our proposed method remains computationally feasible.}

\noindent
\textbf{Source of effectiveness of the KD layer.} To reduce the confounding of factors other than the proposed method, we examine whether the performance increase is coming from the method or the capacity increase introduced by our KD layer. Hence, in \cref{tab:fitnet}, we compare the performance of the three methods as FitNet, FitNet equipped with our KD layer at the penultimate layer with and without supervision. The experiments are done using CIFAR-100 with the teacher-student pair selected as RN110-RN32 and they are averaged over 5 trials. We trained FitNet using the stage-2 outputs and included our KD layer at the output of the penultimate layer. We highlight the inclusion of supervision (\ie distillation loss) using the notations "with" and "without". These results show that even though the capacity of the student is marginally increased due to our KD layer, the major contribution to the performance occurs upon combining it with our supervision.
\input{supplementary/figures/table_Fitnet_RN110_RN32}
% \vspace{-1.4\intextsep}
\input{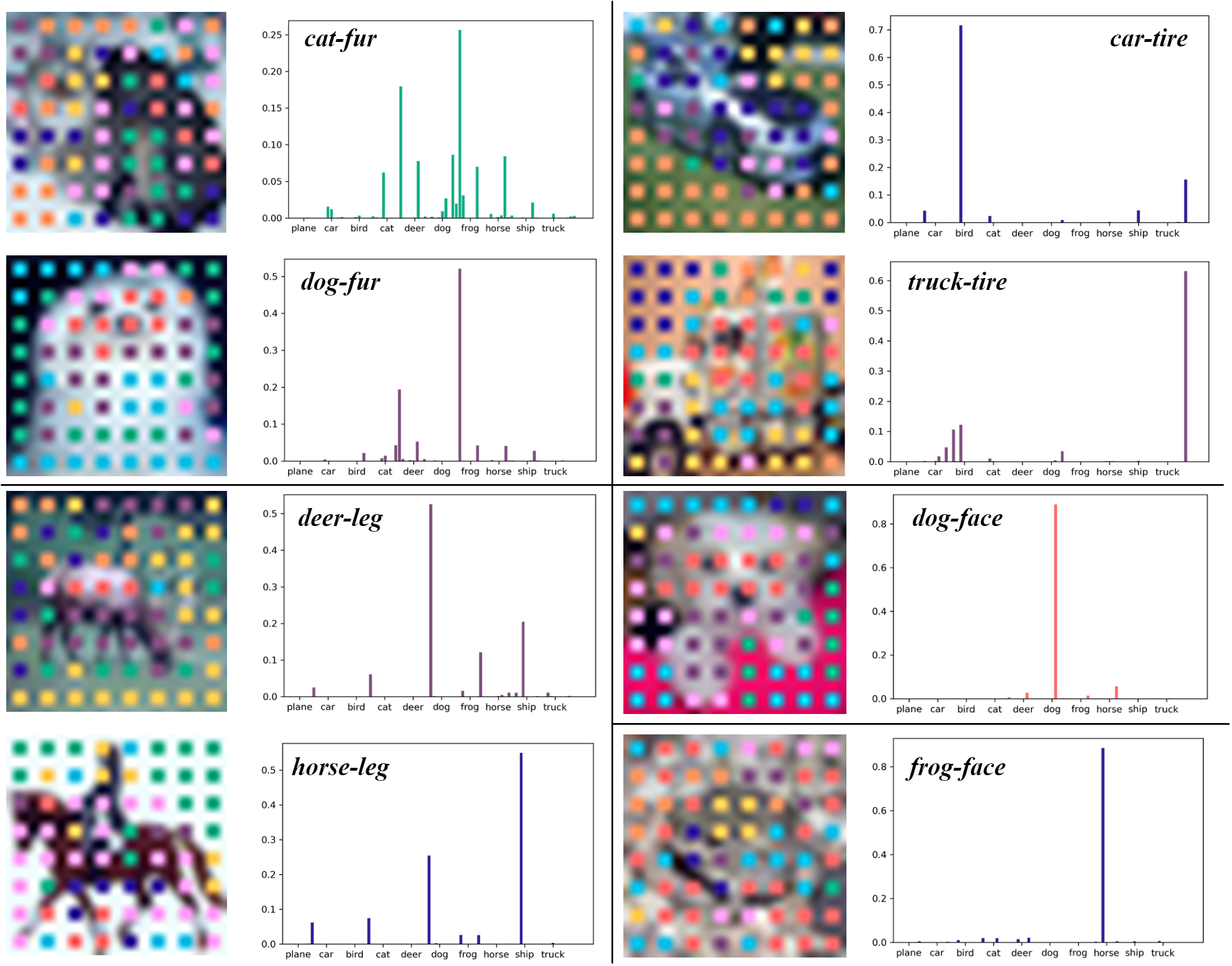}

\noindent
\textbf{Transferred knowledge.} To support the effectiveness of the proposed intermediate layer supervision, we demonstrate the extracted information from the teacher on inter-category relations between the different sub-classes and category-specific patterns for the images in \cref{fig:int_knowledge} using CIFAR-10 with RN56. We mark the center of each patch with respect to its sub-class (denoted by its color). We also provide annotation of a sub-class ($p_{\mathcal{T}}(i)= p(h_2(z_i)\mid h_1(z_i, y))$ in \S~4.2.2 in the main paper) on the right, which shows how discriminating that patch for the teacher for the main task. We observe shared entities such as \emph{tire} for \emph{truck} and \emph{car}, \emph{leg} for \emph{horse} and \emph{deer}, \emph{fur} for \emph{cat} and \emph{dog}. We are also able to observe discriminative patterns such as \emph{face} of a \emph{cat} and a \emph{dog}. Through exploiting this information, the templates of the student (\ie $1{\sxtimes}1$ kernels) adapt to those patterns that the teacher finds useful to discriminate categories. Besides uniquely differentiating discriminative patterns, the student also learns to acknowledge the shared entities by combining them with their learned semantic features. That way, the uninformative patterns such as \emph{fur} that have less peaky distribution can be exploited to represent coarse categories, \eg \emph{belonging to animals}, or can be completely discarded by the student if their matching scores drop below the average due to following an almost uniform distribution.

\section{Empirical Study Details}
% \yetinote{bu kisim extended deneylerden sonraya konabilir.}
In the following sections, we detail our experimental setup, including the utilized datasets, and fully disclose our implementation specifics.

\subsection{Reproducibility}
We provide full details of our experimental setup and recapitulate the implementation details for the sake of complete transparency and reproducibility. Code is available at: \href{\codeurl}{letKD Framework}

\subsection{Experimental Setup}
\label{sec:experiment_details}

\textbf{Datasets.} We adopted three benchmark image classification datasets to extensively evaluate our method, which are also widely used in KD literature. These datasets include CIFAR-100 \cite{krizhevsky2009learning}, which contains 100 classes with 50K training images and 10K test images of size 32x32; Tiny-ImageNet \cite{Le2015TinyIV}, which contains 200 classes with 500 training images, 25 validation images and 25 test images of size 64x64 for each class; ImageNet \cite{imagenet}, which consists over 1.2 million images for training and 50K images for validation which are distributed over 1000 classes. For data augmentation, we use standard operations including normalization that are commonly used in other KD algorithms \cite{jain2019quest}. 
\vspace{\intextsep}

\noindent 
\textbf{Implementation Details.} We perform our experiments on various architectures including ResNet (RN) \cite{HeResnetv1_2015,he2016identity}, Wide ResNet (WRN) \cite{zagoruyko2016wide}, MobileNet (MNV2) \cite{MobileNetV2} and ShuffleNet (SNV1/V2) \cite{ShufflenetV1,ShufflenetV2}. We adopt the framework implemented by \cite{jain2019quest} in PyTorch \cite{NEURIPS2019_9015} to make a fair and unbiased evaluation of our method as well as comparisons with the other invented methods. For the results in \cref{tab:cifar100_sup,tab:imagenet_sup,tab:tinyimagenet}, we use the results of the relevant methods from \cite{jain2019quest,Song_2021_CVPR,Zhao_2022_CVPR}. For SimKD \cite{Chen_2022_CVPR}, we use their official implementation to obtain the corresponding results.

Specifically, for all datasets, we adopt the SGD optimizer with 0.9 Nesterov momentum. For CIFAR-100 and Tiny-ImageNet, we trained for 240 epochs in which the learning rate is divided by 10 at 150th, 180th and 210th epochs. For heterogeneous (MNV2 and SNV1/V2 as students) trainings, we set the initial learning rate as 0.01 and for other architectures, we set the initial learning rate as 0.05. 

For all evaluations on ImageNet, we trained for 100 epochs with an initial learning rate 0.1, which is divided by 10 at 30th, 60th and 90th epochs.

\noindent 
\textbf{Formulation of 1x1 convolution.} In the proposed KD layer including $1\sxtimes1$-BN-ReLU-$1\sxtimes1$ block, we employ $1\sxtimes1$ operations as \textit{normalized} $1\sxtimes1$ \textit{convolutions} with a learnable scale, \ie the kernels are $\ell2$ normalized and scaled. For us, the two \textit{normalized convolutions} serve distinct purposes. For the first $1\sxtimes1$, we utilize this by first normalizing our input features and kernels since we want to measure the cosine similarity between them (\ie their alignment). For the second $1\sxtimes1$, we try to decrease the dependency of the hyperparameter $\lbrace \alpha_{inter}, \alpha_{penult} \rbrace$ through the normalized and scaled outputs to have a more stable learning. Owing to this mechanism, we ease the process of adding the importance and attention gathered from $1\sxtimes1$-BN-ReLU-$1\sxtimes1$ to the features at the shortcut connection.
\vspace{.5\intextsep}

\noindent 
\textbf{Heterogeneous distillation cases.}
For heterogeneous cases, we transform the features of the teacher before obtaining the soft assignments by applying an additional average pooling operation before quantization (\emph{K-means} operation) to align its spatial dimensions with the predictions of the student. For instance, at the penultimate layer, RN32x4 and WRN-40-2 have $8{\sxtimes}8$ feature maps, SNV1/V2 and RN50 have $4{\sxtimes}4$ feature maps, and MNV2 has $2{\sxtimes}2$ feature maps. Hence, the spatial dimensions between the teacher and the student should be aligned to apply the distillation loss properly.

\section{Implementations with Pseudo-codes}
\label{sec:sub_algos}
\input{supplementary/algo_teacher_penultimate}
\input{supplementary/algo_teacher_intermediate}
\input{supplementary/algo_student}

%% file: supplementary/figures/table_CIFAR100.tex
% \usepackage{graphicx}

\begin{table*}[ht]%[13]{r}[-5pt]{0.6\linewidth}
    %\vspace{-.9\intextsep}
	\centering
	\caption{Average top-1 accuracies on CIFAR-100 over 5 trials. \textbf{Bold}: best in its category.}
	\label{tab:cifar100_sup}
	\resizebox{\linewidth}{!}{%
    	\begin{tabular}{cccccccccccc}
    		\toprule 
    		Archs. $\rightarrow$ & \multicolumn{6}{c}{\textbf{Homogeneous}} & & \multicolumn{4}{c}{\textbf{Heterogeneous}} \\ \cmidrule{2-7} \cmidrule{9-12} 
    		Teacher & WRN-40-2 & WRN-40-2 & RN56 & RN110 & RN110 & RN32x4 & & WRN-40-2 & RN32x4 & RN32x4 & RN50\\  
    		Student & WRN-16-2 & WRN-40-1 & RN20 & RN20  & RN32  & RN8x4 & & SNV1 & SNV1 & SNV2 & MNV2\\ \midrule
    		\multirow{2}{*}{Methods $\downarrow$} & 75.61 & 75.61 & 72.34 & 74.31 & 74.31 & 79.42 & & 75.61 & 79.42 & 79.42 & 79.34 \\
    		 & 73.26 & 71.98 & 69.06 & 69.06 & 71.14 & 72.50 & & 70.50 & 70.50 & 71.82 & 64.60 \\ \midrule
            KD & 74.92 & 73.54 & 70.66 & 70.67 & 73.08 & 73.33 & & 74.83 & 74.07 & 74.45 & 67.35\\
            FitNet & 73.58 & 72.24 & 69.21 & 68.99 & 71.06 & 73.50 & & 73.73 & 73.59 & 73.54 & 63.16 \\
            AT & 74.08 & 72.77 & 70.55 & 70.22 & 72.31 & 73.44 & & 73.32 & 71.73 & 72.73 & 58.58 \\
            AB & 72.50 & 72.38 & 69.47 & 69.53 & 70.98 & 73.17 & & 73.34 & 73.55 & 74.31 & 67.20 \\
            FSP & 72.91 & - & 69.95 & 70.11 & 71.89 & 72.62 & & - & - & - & - \\
            SP & 73.83 & 72.43 & 69.67 & 70.04 & 72.69 & 72.94 & & 74.52 & 73.48 & 74.56 & 68.08 \\
            VID & 74.11 & 73.30 & 70.38 & 70.16 & 72.61 & 73.09 & & 73.61 & 73.38 & 73.40 & 67.57 \\
            CRD & 75.48 & 74.14 & 71.16 & 71.46 & 73.48 & 75.51 & & 76.05 & 75.11 & 75.65 & 69.11 \\
            $\underset{\text{+KD}}{\text{CRD}}$ & 75.64 & 74.38 & 71.63 & 71.56 & 73.75 & 75.46 & & 76.27 & 75.12 & 76.05 & 69.54\\
            DKD & 76.24 & 74.81 & 71.97 & - & 74.11 & 76.32 & & 76.70 & 76.45 & 77.07 & 70.35 \\
            SimKD & 76.06 & 74.92 & 68.95 & 69.35 & 72.15 & \textbf{78.08} & & 76.95 & 77.18 & 77.78 & 68.91 \\
            TDD & 75.01 & 74.04 & 71.53 & - & - & - & & 75.60 & - & - & 68.37 \\
            $\underset{\text{+CRD}}{\text{TDD}}$ & 75.71 & 74.35 & 71.88 & - & - & - & & 76.34 & - & - & 69.22 \\
            QUEST & 76.10 & 74.58 & 71.84 & 71.89 & 74.08 & 75.88 & & 76.75 & 76.28 & 77.09 & 69.81 \\
            \midrule
            \textbf{letKD-1} & 
            $\underset{\scriptsize{\mp0.15}}{76.29}$ & $\underset{\scriptsize{\mp0.09}}{75.01}$ & $\underset{\scriptsize{\mp0.24}}{72.44}$ & $\underset{\scriptsize{\mp0.31}}{72.68}$ & $\underset{\scriptsize{\mp0.14}}{74.40}$ & $\underset{\scriptsize{\mp0.06}}{76.70}$ & & $\underset{\scriptsize{\mp0.16}}{76.93}$ & $\underset{\scriptsize{\mp0.24}}{76.65}$ & $\underset{\scriptsize{\mp0.17}}{77.75}$ &
            $\underset{\scriptsize{\mp0.18}}{69.97}$ \\
            \textbf{letKD-2} & 
    $\mathbf{\underset{\scriptsize{\mp0.22}}{76.56}}$ & $\mathbf{\underset{\scriptsize{\mp0.13}}{75.19}}$ & $\mathbf{\underset{\scriptsize{\mp0.16}}{73.27}}$ & $\mathbf{\underset{\scriptsize{\mp0.14}}{73.38}}$ & $\mathbf{\underset{\scriptsize{\mp0.20}}{74.62}}$ & $\mathbf{\underset{\scriptsize{\mp0.18}}{77.09}}$ & & $\mathbf{\underset{\scriptsize{\mp0.12}}{77.08}}$ & $\mathbf{\underset{\scriptsize{\mp0.12}}{77.30}}$ & $\mathbf{\underset{\scriptsize{\mp0.06}}{77.95}}$ &
    $\mathbf{\underset{\scriptsize{\mp0.23}}{70.39}}$ \\
    		\bottomrule
    	\end{tabular}%
    }
\end{table*}

% \textbf{$\pm$0.1}

%% file: supplementary/figures/table_ImageNet.tex
% \usepackage{graphicx}

\begin{table*}[ht]%[13]{r}[-5pt]{0.6\linewidth}
    %\vspace{-.9\intextsep}
	\centering
	\caption{Top-1 and top-5 accuracies on ImageNet. Setting \textbf{(a)}: Teacher and student models are selected as RN34-RN18. Setting \textbf{(b)}: Teacher and student models are selected as RN50-MNV2. \textbf{Bold}: best in its category.}
	\label{tab:imagenet_sup}
        \resizebox{\textwidth}{!}{
	\begin{tabular}{c|ccccccccc}
		\toprule 
		Setting & & Teacher & Student & KD & AT+KD & DKD & QUEST & \textbf{letKD-1} & \textbf{letKD-2} \\ \midrule   
		\multirow{2}{*}{(a)} & Top-1 & 73.31 & 69.75 & 70.66 & 70.70 & 71.70 & 71.67 & 72.33 & \textbf{72.38} \\
		& Top-5 & 91.42 & 89.07 & 89.88 & 90.00 & 90.41 & 90.67 & 91.06 & \textbf{91.15}\\ \midrule
		\multirow{2}{*}{(b)} & Top-1 & 76.13 & 68.87 & 68.58 & 69.56 & 72.05 & 72.54 & 73.78 & \textbf{73.98}\\
		& Top-5 & 92.86 & 88.76 & 88.98 & 89.33 & 91.05 & 91.13 & 91.81 & \textbf{92.00}\\
		\bottomrule
	\end{tabular}%
        }
\end{table*}

%% file: supplementary/figures/table_TinyImageNet.tex
% \usepackage{graphicx}

\begin{table}[H]%[13]{r}[-5pt]{0.6\linewidth}
% \begin{wraptable}[15]{r}[-5pt]{0.5\linewidth}
    % \vspace{-.9\intextsep}
    % \vspace{-10}
	\centering
	\caption{Average top-1 accuracies on Tiny-ImageNet over 3 trials. \textbf{Bold}: best in its category.}
	\label{tab:tinyimagenet}
	\resizebox{0.68\linewidth}{!}{%
	\begin{tabular}{ccccccc}
		\toprule 
		Archs. $\rightarrow$ & \multicolumn{3}{c}{\textbf{Homogeneous}} & & \multicolumn{2}{c}{\textbf{Heterogeneous}} \\ \cmidrule{2-4} \cmidrule{6-7}
		Teacher & WRN-40-2 & WRN-40-2 & RN56 & & WRN-40-2 & RN50 \\  
		Student & WRN-16-2 & WRN-40-1 & RN20 & & SNV1  & MNV2 \\ \midrule
		\multirow{2}{*}{Methods $\downarrow$} & 61.26 & 61.26 & 58.34 & & 61.26 & 68.97 \\
		& 57.17 & 56.25 & 52.66 & & 60.52 & 58.35 \\ \midrule
        KD & 59.16 & 57.75 & 53.04 & & 64.80 & 58.68 \\
        FitNet & 57.75 & - & 51.73 & & - & 57.55 \\
        AT & 58.71 & 57.41 & 54.01 & & 63.90 & 50.91\\
        FSP & 57.33 & - & 53.55 & & - & - \\
        SP & 55.69 & 53.74 & 54.03 & & 64.62 & 58.11 \\
        VID & 58.51 & 57.45 & 53.20 & & 63.58 & 57.50 \\
        TDD & 59.22 & 58.42 & 54.45 & & 65.27 & 59.09 \\
        $\underset{\text{+CRD}}{\text{TDD}}$ & 59.53 & 59.20 & 54.85 & & 65.50 & 59.72\\
        QUEST & 59.86 & 59.13 & 54.53 & & 65.23 & 59.81 \\
        \midrule
        \textbf{letKD-1} & $\underset{\scriptsize{\mp0.36}}{61.42}$ & $\underset{\scriptsize{\mp0.50}}{59.75}$ & $\underset{\scriptsize{\mp0.33}}{55.54}$ & & $\underset{\scriptsize{\mp0.42}}{65.70}$ & $\underset{\scriptsize{\mp0.17}}{60.69}$ \\
        \textbf{letKD-2} & $\mathbf{\underset{\scriptsize{\mp0.17}}{62.21}}$ & $\mathbf{\underset{\scriptsize{\mp0.27}}{60.59}}$ & $\mathbf{\underset{\scriptsize{\mp0.46}}{57.35}}$ & & $\mathbf{\underset{\scriptsize{\mp0.50}}{66.15}}$ & $\mathbf{\underset{\scriptsize{\mp0.46}}{61.15}}$ \\
		\bottomrule
	\end{tabular}%
	}
\end{table}

%% file: supplementary/figures/fig_subclass_number.tex
% Below is an example of how to insert images. Delete the ``\vspace'' line,
% uncomment the preceding line ``\centerline...'' and replace ``imageX.ps''
% with a suitable PostScript file name.
% -------------------------------------------------------------------------
\begin{wrapfigure}[12]{l}[-2pt]{0.45\linewidth}
\vspace{-1\intextsep}
%\begin{minipage}{.49\linewidth}
%\begin{figure}[!ht]
  %\centering
  \centerline{\includegraphics[width=1.0\linewidth,keepaspectratio]{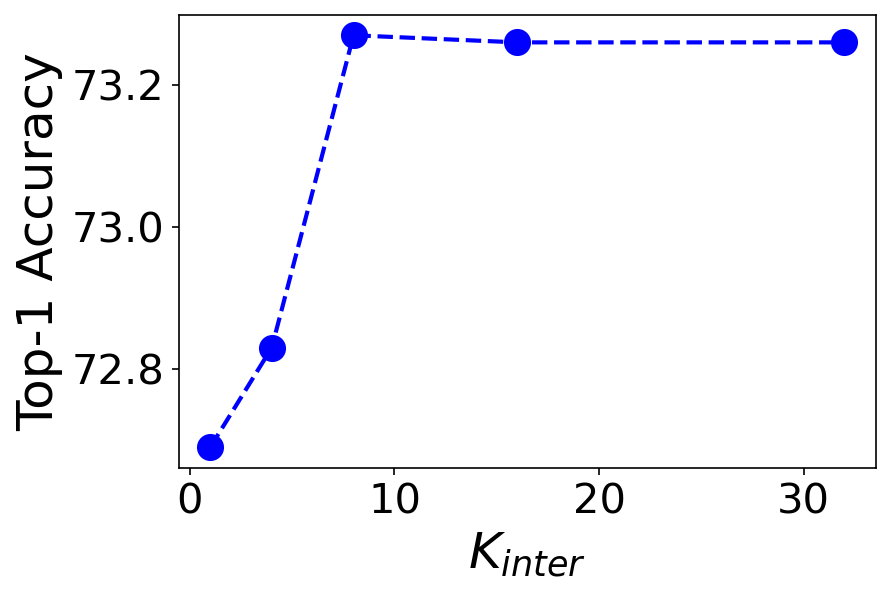}}
  %\smallskip
  %\centerline{(b)}
  %\end{figure}
%\end{minipage}
\caption{Effect of $K_{inter}$}
\label{fig:subclass_number}
\end{wrapfigure}

%% file: supplementary/figures/table_subclass_value_offline_effect.tex
\begin{table}[h]
	\centering
	\caption{Effect of our KD layer on intermediate layer classification performance on CIFAR-100 with RN56-RN20}
        \vspace{.2\intextsep}
	\label{tab:subclass_value_offline}
        \resizebox{0.5\linewidth}{!}{
    	\begin{tabular}{ccc}
    	\toprule
    	\multirow{2}{*}{letKD-2 ($\alpha_{inter}=0$)} & \multicolumn{2}{c}{letKD-2 ($\alpha_{inter}=1$)} \\ \cmidrule{2-3}
        & $x$ & $\hat{x}$ \\ \midrule
        52.71 & 51.71 & 56.36 \\ \bottomrule
    	\end{tabular}%
     }
% \end{table*}
\end{table}

%% file: supplementary/figures/table_value_effect.tex
\begin{wraptable}[11]{r}[-5pt]{0.5\linewidth}
\vspace{-.9\intextsep}
\centering
\caption{Effect of the included parts in letKD-2}
\label{tab:value_effect}
    \resizebox{\linewidth}{!}{%
        \begin{tabular}{ccccc}
            \toprule 
            Inter. & $\alpha_{inter}$ & Penult. & $\alpha_{penult}$ & Top-1 Acc. \\ \midrule
            \checkmark & 0 & - & 0 & 70.64 \\ 
            \checkmark & 1 & - & 0 & 70.80 \\
            - & 0 & \checkmark & 0 & 71.84 \\  
            - & 0 & \checkmark & 1 & 72.44 \\   
            \checkmark & 0 & \checkmark & 0 & 71.70 \\  
            \checkmark & 0 & \checkmark & 1 & 72.13 \\  
            \checkmark & 1 & \checkmark & 0 & 72.78 \\  
            \checkmark & 1 & \checkmark & 1 & 73.27 \\ 
            \bottomrule
        \end{tabular}
    }
% \end{table}
\end{wraptable}

%% file: supplementary/figures/table_Fitnet_RN110_RN32.tex
% % \begin{wraptable}[8]{r}[-3pt]{0.6\linewidth}
% \begin{table}[h]
%     % \vspace{-1.5\intextsep}	

%     \centering
%     \caption{Effect of the impact of the KD layer on performance improvement due to capacity increase}
% 	\label{tab:fitnet}
% 	\resizebox{0.7\linewidth}{!}{%
%     	\begin{tabular}{cccccc}
%     		\toprule 
%     		 \multicolumn{2}{c}{Stage-2 Output} & & \multicolumn{2}{c}{Stage-3 Output} & \multirow{2}{*}{Top-1 Acc.}\\ \cmidrule{0-1} \cmidrule{4-5} 
%              $\alpha_{inter}$ & Distill Loss & & $\alpha_{penult}$ & Distill Loss & \\ \midrule
%              0 & FitNet & & 0 & - & 71.59 \\
%              0 & FitNet & & 1 & - & 71.80 \\
%              0 & FitNet & & 1 & QUEST & 73.36 \\
%              \bottomrule		
%     	\end{tabular}%
%     }
% \end{table}

% \begin{wraptable}[8]{r}[-3pt]{0.6\linewidth}
\begin{table}[H]
    % \vspace{-1.5\intextsep}	

    \centering
    \caption{Effect of the impact of the KD layer on performance improvement considering the capacity increase}
	\label{tab:fitnet}
	\resizebox{0.6\linewidth}{!}{%
    	\begin{tabular}{cc}
    		\toprule 
             Methods & Top-1 Acc \\ \midrule
             FitNet & 71.59 \\
             FitNet+KD layer without supervision & 71.80 \\
             FitNet+KD layer with supervision & 73.36 \\
             \bottomrule		
    	\end{tabular}%
    }
\end{table}

%% file: supplementary/figures/fig_int_knowledge.tex
\begin{figure*}[!ht]
      \centering
      \centerline{\includegraphics[width=\linewidth]{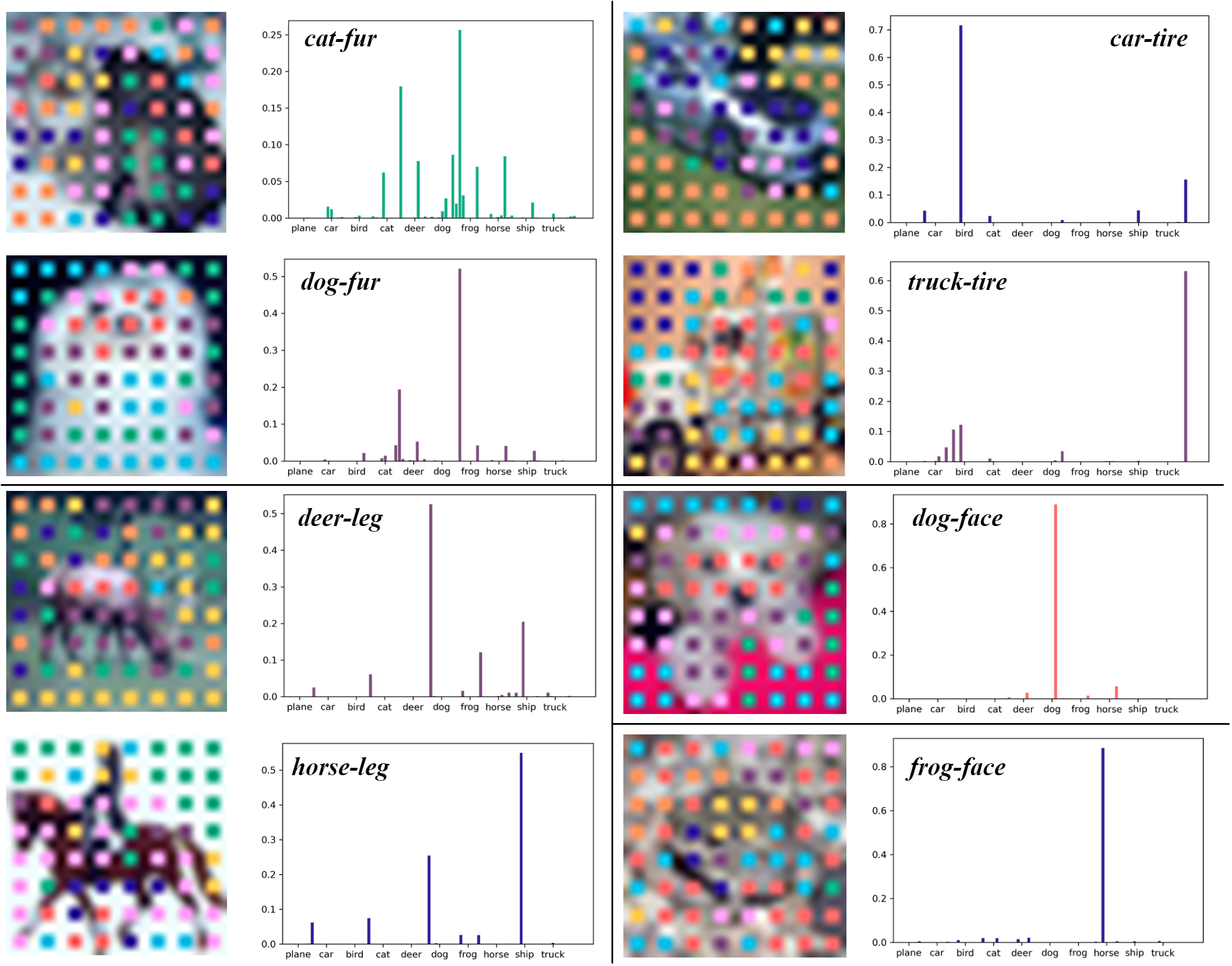}}
    %  \vspace{2.0cm}
  \caption{Sample images with the teacher sub-class annotations marking the center of each spatial location. Each color corresponds to a distinct sub-class. On the right of each image, the histogram for the sub-class assignments are plotted, where $x$-axis corresponds to subclass indices. The indices  of the sub-classes associated with its super-class lie on the right of the index ticked by the class label.
  }
	\label{fig:int_knowledge}
  \end{figure*}

%% file: supplementary/algo_teacher_penultimate.tex
\renewcommand{\algorithmiccomment}[1]{\hfill\eqparbox{COMMENT}{\footnotesize{//#1}}}
\renewcommand{\algorithmicrequire}{\textbf{input:}}
\renewcommand{\algorithmicensure}{\textbf{repeat}}
\newcommand{\algrule}[1][.2pt]{\par\vskip.5\baselineskip\hrule height #1\par\vskip.5\baselineskip}

\begin{algorithm}[h]
\caption{TEACHER PENULTIMATE LAYER KD}\label{algo:algo_penultimate}
\textbf{offline:}
\algrule
\begin{algorithmic}
\Require $X = \lbrace x_i\rbrace_{i\in [\mathcal{X}_T]}$, $K_{penult}$, $\theta_t$   \algorithmiccomment{all training images, \# of cluster centers,}
% \vspace{-.8\intextsep}
\State  \algorithmiccomment{parameters of the teacher}
\State $F \gets f_t^{(\shortminus 1)}(X; \theta_t)$ \algorithmiccomment{teacher's features at the penultimate layer,}
\State \algorithmiccomment{$F \in \mathbb{R}^{[\mathcal{X}_T].w.h \sxtimes d}$}
\State $\lbrace \rho_k \rbrace _{k \in [K_{penult}]} \gets \mathrm{KMeans}(F, K_{penult})$ \algorithmiccomment{clustering operation to the teacher's features}
\end{algorithmic}
\textbf{return} $\lbrace \rho_k \rbrace _{k \in [K_{penult}]}$ \algorithmiccomment{cluster centers in the quantized space}
\algrule[.5pt]
\textbf{online:} 
\algrule
\begin{algorithmic}
\Require $X = \lbrace x_i\rbrace_{i\in [b]}, \lbrace \rho_k \rbrace _{k \in [K_{penult}]}, \theta_t$ \algorithmiccomment{batch of images, cluster centers, parameters}
% \vspace{-.8\intextsep}
\State \algorithmiccomment{of the teacher}
\State $F \gets f_t^{(\shortminus 1)}(X; \theta_t)$ \algorithmiccomment{teacher's features at the penultimate layer,}
\State \algorithmiccomment{$F \in \mathbb{R}^{[b].w.h \sxtimes d}$}
\State $d \gets [\Vert F - \rho_k\Vert_2^2]_{k\in[K_{penult}]}$  \algorithmiccomment{distance of each spatial location to the}
% \vspace{-.8\intextsep}
\State \algorithmiccomment{cluster centers}
\State $p_\mathcal{T} \gets \mathrm{softmax}(d)$ \algorithmiccomment{soft assignments of the teacher for each}
% \vspace{-.8\intextsep}
\State \algorithmiccomment{spatial location}
\end{algorithmic}
\textbf{return} $p_\mathcal{T}$ \algorithmiccomment{soft assignments of the teacher}
\end{algorithm}

%% file: supplementary/algo_teacher_intermediate.tex
\renewcommand{\algorithmiccomment}[1]{\hfill\eqparbox{COMMENT}{\footnotesize{//#1}}}

\begin{algorithm}[H]
\caption{TEACHER INTERMEDIATE LAYER KD}\label{algo:algo_intermediate}
\textbf{offline:}
\algrule
\begin{algorithmic}
\Require $(X,Y) = (\lbrace x_i\rbrace, \lbrace y_i\rbrace)_{i\in [\mathcal{X}_T]}$, $K_{inter}$, $\theta_t$, $C$   
% \vspace{-.8\intextsep}
\State \algorithmiccomment{all training image-label  pairs, \# of sub-classes,}
\State \algorithmiccomment{parameters of the teacher, \# of classes}
\State $F \gets f_t^{(l^\prime)}(X; \theta_t)$ \algorithmiccomment{teacher features at $l^\prime$th layer}
\State $W_\mathrm{LDA}, b_\mathrm{LDA} \gets \mathrm{LDA}(F, Y)$ \algorithmiccomment{obtain weight and bias for LDA} 
\State $F_\mathrm{LDA} \gets \mathrm{Conv}1\sxtimes1(F, W_\mathrm{LDA}, b_\mathrm{LDA})$ \algorithmiccomment{apply LDA, $F_\mathrm{LDA} \in \mathbb{R}^{[b].w.h \sxtimes d_\mathrm{LDA}}$}
\For{$c=1:C$} 
    \State $F_\mathrm{LDA}^c \gets F_\mathrm{LDA}[y=c]$ \algorithmiccomment{obtain the LDA features belonging to class $c$}
    \State $\rho_c, P^c \gets \mathrm{KMeans}(F_\mathrm{LDA}^c, K_{inter})$ \algorithmiccomment{obtain clusters and predictions through}
    \State \algorithmiccomment{$K$-means clustering}
    \For{$k=1:K_{inter}$} 
        \State $F_\mathrm{LDA}^{k,c} \gets F_\mathrm{LDA}[P^c=k]$ \algorithmiccomment{obtain LDA features belonging to sub-class $k$}
        \State \algorithmiccomment{in class $c$}
        \State $\mathrm{prot}^{k,c} \gets \mathrm{mean}(F_\mathrm{LDA}^{k,c})$ \algorithmiccomment{obtain representative prototype for sub-class}
        \State \algorithmiccomment{$k$ in class $c$}
    \EndFor
\EndFor
\State $\mathrm{prot} \gets \lbrace \mathrm{prot}^{k,c}\rbrace_{k\in [K_{inter}]}^{c\in [C]}$
\State $\lbrace s_\mathcal{T}^{k,c}\rbrace_{k\in [K_{inter}]}^{c \in [C]} \gets \mathrm{NNSearch}(F_\mathrm{LDA}, \mathrm{prot})$ 
\State \algorithmiccomment{apply NN Search between features and}
\State \algorithmiccomment{prototypes}
\State $s_\mathcal{T} \gets \lbrace s_\mathcal{T}^{k,c}\rbrace_{k\in [K_{inter}]}^{c \in [C]}$ \algorithmiccomment{scores for all sub-classes,}
\State \algorithmiccomment{$s_\mathcal{T} \in \mathbb{R}^{K_{inter}.C \sxtimes K_{inter}.C}$}
\end{algorithmic}
\textbf{return} $s_\mathcal{T}, W_\mathrm{LDA}, b_\mathrm{LDA}, \lbrace \rho_c^k \rbrace_{c \in [C]}^{k \in [K_{inter}]}$ \algorithmiccomment{scores, LDA parameters, cluster centers}
\algrule[.5pt]
\textbf{online:} 
\algrule
\begin{algorithmic}
\Require $(X,Y) = (\lbrace x_i\rbrace, \lbrace y_i\rbrace)_{i\in [b]}, s_\mathcal{T}, W_\mathrm{LDA}, b_\mathrm{LDA}, \lbrace \rho_c^k \rbrace_{c \in [C]}^{k \in [K_{inter}]}, \theta_t$ 
\State \algorithmiccomment{batch of image-label pairs, scores, LDA}
\State \algorithmiccomment{parameters, cluster centers, parameters of the}
\State \algorithmiccomment{teacher}
\State $F_\mathrm{LDA} \gets \mathrm{Conv}1\sxtimes1(f_t^{(l^\prime)}(X;\theta_t), W_\mathrm{LDA}, b_\mathrm{LDA})$ 
\State \algorithmiccomment{apply LDA to the teacher's features at $l^\prime$th layer}
\State \algorithmiccomment{$F_\mathrm{LDA} \in \mathbb{R}^{[b].w.h \sxtimes d_\mathrm{LDA}}$}
\State $k^{*} \gets \mathrm{min}([\Vert F_\mathrm{LDA} - \rho_Y^k  \Vert_2^2]_{k \in [K_{inter}]})$ \algorithmiccomment{assign the closest sub-class cluster index for}
\State \algorithmiccomment{class $Y$ to each spatial location}
\State $p_\mathcal{T} \gets s_\mathcal{T}(k^{*})$ \algorithmiccomment{assign the score of the selected cluster}
\end{algorithmic}
\textbf{return} $p_\mathcal{T}$ \algorithmiccomment{soft assignments of the teacher}
\end{algorithm}

%% file: supplementary/algo_student.tex
\renewcommand{\algorithmiccomment}[1]{\hfill\eqparbox{COMMENT}{\footnotesize{//#1}}}

% KD Student
\begin{algorithm}[H]
\caption{STUDENT KD SUPERVISION}\label{algo:algo_student}
\textbf{online:}
\algrule
\begin{algorithmic}
\Require $X = \lbrace x_i\rbrace_{i\in [b]}$, $p_\mathcal{T}$, $\theta_s$ \algorithmiccomment{batch of images, soft assignments of the teacher,}
\State \algorithmiccomment{parameters of the student}
\State $F \gets f_s^{(l)}(X; \theta_s)$ \algorithmiccomment{student's features at the $l$th (penultimate or}
\State \algorithmiccomment{intermediate) layer}
\State $\hat{F},p_\mathcal{S} \gets \mathrm{KDLayer}(F;\theta_s)$ \algorithmiccomment{see Fig. 2 in the main paper}
\State $\mathcal{L}_{KD} \gets \mathrm{KLDiv}(p_\mathcal{T}, p_\mathcal{S})$ \algorithmiccomment{distillation loss using teacher's soft assignments}
\State \algorithmiccomment{and student's predictions}
\end{algorithmic}
\textbf{return} $\hat{F},\mathcal{L}_{KD}$ \algorithmiccomment{output features of the KD layer, distillation loss}
\end{algorithm}

%% file: supplementary/appendix.tex
\renewcommand{\theequation}{A.\arabic{equation}}
\section*{Appendix}
\subsection*{BN-ReLU as a Soft Maximizer}
\label{sec:bnargmax}
% \yetinote{Burada 3x3 uzerinden anlatmak kafa karistirici olabilir. Paperda formulastonlar hep 1x1 uzerinden.}
To strengthen our approximation of BN-ReLU as a soft maximizer considering the \hypertarget{argmax_sup}{problem}:
\begin{equation}
p_{\mid i} =  \argmax_{p,q\geqslant 0} q\,\mu+\Sigma_k p_k\, \omega_k\T x_i \quad \text{s.to} \quad q+\Sigma_k p_k =1
\end{equation}
we start with the explanation of the overall process, where $x_i$ represents a local region $i$ in a feature map $x$, $\{\omega_k\in\R^{d}\}_{k\in[K]}$ are the $1\sxtimes1$ kernels, and $\mu$ is a threshold enabling to zero out the embedding vector if no kernel matches with at least $\mu$ similarity. As we state in the paper, we make this problem differentiable by employing entropy smoothing to the problem in (\hyperlink{argmax_sup}{A.1}) \hypertarget{entropy_argmax_sup}{as}:
\begin{equation}
p_{\mid i} = \argmax_{p,q\geqslant0} q\,\mu+ p\T a_{\mid i} -\tfrac{1}{\epsilon}(q\log q+p\T\log p) \quad \text{s.to} \quad q+\Sigma_k p_k =1
\end{equation}
and obtain a soft-max solution:
\begin{equation}\label{eq:softmax}
p_{k\mid i} = \tfrac{\exp(\epsilon a_{k\mid i})}{\exp(\epsilon \mu)+\Sigma_{k^\prime}\exp(\epsilon a_{k^\prime\mid i})}
\end{equation}
where $a_{k\mid i} = \omega_k\T x_i $ and $\epsilon$ controls the smoothness of $p_{\mid i}$. As can be seen from \eqref{eq:softmax}, apart from the temperature parameter $\epsilon$, we only need to add an additional dimension with the value $\mu$ to the channels of $a_{\mid i}$ to mimick the threshold in (\hyperlink{argmax_sup}{A.1}). Yet, the problem here is finding proper $\mu$ and $\epsilon$ values. Indeed, BN-ReLU is shown to mitigate that problem in \cite{Gorgun_2022_BMVC} and the equivalence of the solution in \eqref{eq:softmax} and BN-ReLU (up to a scale) is empirically validated.

We now derive an alternative equivalence to rather explicitly show that replacing soft-max with BN-ReLU inherently makes the model learn these parameters while performing a scaled version of soft-max.

Note that BN \cite{normalization2015accelerating} and its successor counterparts \cite{ulyanov2016instance,ba2016layer,wu2018group} perform activity normalization using some batch statistics as:
\begin{equation}\label{eq:BN}
\mathrm{BN}(a_k)=\gamma_k\tfrac{a_k-\mathbb{E}[a_k]}{\sqrt{\mathrm{Var}(a_k)}}+\beta_k=\gamma_k\hat{a}_k + \beta_k 
\end{equation}
which can be interpreted as \textit{whitening} its input with a learnable scale and bias, where $\mathbb{E}[a_k]$ and $\mathrm{Var}[a_k]$ are calculated using the whole batch.

To employ BN-ReLU as a replacement of (\hyperlink{entropy_argmax_sup}{A.2}), we first consider the whitened version of our activations, \ie $\hat{a}_k = \frac{a_k - \mathbb{E}(a_k)}{\sqrt{\mathrm{Var}(a_k)}}$, to be used in \eqref{eq:softmax} and use a scale $\gamma_k$ to make their values around 0 as $a_k^\prime = \gamma_k \hat{a}_k$. Then, when we apply soft-max to $a_{k \mid i}^\prime$, we can use the first order Taylor series expansion to approximate the unnormalized soft-max operation applied to them as:
\begin{equation}\label{eq:taylor_series}
e^{a_{k \mid i}^\prime} \approx 1 + a_{k \mid i}^\prime + \frac{a_{k \mid i}^{\prime^2}}{2!} + \frac{a_{k \mid i}^{\prime^3}}{3!} + ... \approx 1 + a_{k \mid i}^\prime + \mathrm{err} 
\end{equation}
where $\mathrm{err}$ is an error owing to the higher order terms. When we consider the expression in \eqref{eq:softmax} with the inclusion of temperature $\epsilon$, we can say that for certain $k$'s, $p_{k\mid i}$ will go to zero if $\exp(\epsilon a_{k\mid i})$ is way smaller than the denominator (sum of all exponential terms including the effect of the threshold parameter $\mu$) of \eqref{eq:softmax}. Moreover, since the soft-max formulation is linearized in \eqref{eq:taylor_series}, the output might be negative. Hence, if we employ the solution in \eqref{eq:taylor_series}, the condition that should be satisfied for $p_{k\mid i}$ to be non-zero and non-negative would be: 
\begin{equation}\label{eq:softmax_th}
1 + a_{k\mid i}^\prime + \mathrm{err} > \mathrm{th} \rightarrow 1 + a_{k\mid i}^\prime + \mathrm{err} - \mathrm{th} > 0
\end{equation}
where the terms $\{1,\mathrm{th},\mathrm{err}\}$ can be combined into a single term $\beta$ (involving the effect of $\mu$) as:
\begin{equation}\label{eq:softmax_beta}
a_{k\mid i}^\prime + \beta_k > 0 \rightarrow \gamma_k \hat{a}_{k\mid i} + \beta_k > 0
\end{equation}
Internally, the constraint defined in \eqref{eq:softmax_beta} mimicks the function ReLU. When this constraint is satisfied, the terms $\gamma_k\hat{a}_{k\mid i} + \beta_k$ of the unnormalized soft-max would be counted as the corresponding outputs. Hence, when we consider the relationship between \eqref{eq:softmax_beta} and \eqref{eq:BN}, if we use BN+ReLU as a replacement of \eqref{eq:softmax}, we can simply employ $\hat{p}_k=\mathrm{max}\{0, \gamma_k\hat{a}_k+\beta_k\}$, \ie $\mathrm{ReLU}$, to zero-out the assignment vector and let BN learn the proper parameters, $(\beta_k,\gamma_k)$, using the batch statistics to assess the poor matching scores. For the pixels with non-zero activations after BN-ReLU, we can obtain the normalized assignment vector as $\nicefrac{\hat{p}_k}{\eta}$ where $\eta\coloneqq\Sigma_k\hat{p}_k$. That being said, we empirically find that, absorbing $\eta$ into $\alpha$ (Eq. 4.1 in the main paper) and $\alpha$ into $\gamma_k$, are useful to adaptively put more emphasis on the teacher's knowledge according to the matching scores.